\documentclass[12pt]{colt2015} % Include author names

\title{A Chaining Algorithm for Online Nonparametric Regression}

\usepackage{times}
\usepackage[utf8]{inputenc}
\usepackage{enumitem}

\allowdisplaybreaks

 \coltauthor{\Name{Pierre Gaillard} \Email{pierre-p.gaillard@edf.fr}\\
 \addr EDF R\&D, Clamart, France \\
GREGHEC: HEC Paris – CNRS, Jouy-en-Josas, France
 \AND
 \Name{S\'{e}bastien Gerchinovitz} \Email{sebastien.gerchinovitz@math.univ-toulouse.fr}\\
 \addr Institut de Mathématiques de Toulouse, Université Paul Sabatier, France
 }

\newenvironment{proofref}[1]{\par\noindent{\bfseries\upshape Proof (of~#1)\ }}{\hfill\BlackBox\\[2mm]}

\renewcommand{\epsilon}{\varepsilon}
\renewcommand{\leq}{\leqslant}
\renewcommand{\geq}{\geqslant}
\renewcommand{\phi}{\varphi}
\renewcommand{\hat}{\widehat}

\newcommand{\R}{\mathbb{R}}
\newcommand{\N}{\mathbb{N}}
\newcommand{\bg}{\boldsymbol{g}}
\newcommand{\bu}{\boldsymbol{u}}
\newcommand{\bw}{\boldsymbol{w}}
\newcommand{\cA}{\mathcal{A}}
\newcommand{\cX}{\mathcal{X}}
\newcommand{\cF}{\mathcal{F}}
\newcommand{\cC}{\mathcal{C}}
\newcommand{\cO}{\mathcal{O}}
\newcommand{\cY}{\mathcal{Y}}
\newcommand{\cB}{\mathcal{B}}
\newcommand{\cG}{\mathcal{G}}
\newcommand{\cN}{\mathcal{N}}

\newcommand{\cP}{\mathcal{P}}
\newcommand{\cQ}{\mathcal{Q}}

\newcommand{\cT}{\mathcal{T}}
\newcommand{\E}{\mathbb{E}}
\newcommand{\indicator}[1]{\mathbb{I}_{#1}}

\newcommand{\bc}{\boldsymbol{c}}

\newcommand{\eqdef}{\triangleq}
\newcommand{\norm}[1]{\left\lVert#1\right\rVert}
\DeclareMathOperator{\Reg}{Reg}
\DeclareMathOperator{\argmin}{argmin}
\DeclareMathOperator{\diam}{\mathrm{diam}}
\DeclareMathOperator{\card}{\mathrm{card}}

\newlength{\minipagewidth}
\begin{document}

\maketitle

\vspace{-0.8cm}
\begin{abstract}
We consider the problem of online nonparametric regression with arbitrary deterministic sequences. Using ideas from the chaining technique, we design an algorithm that achieves a Dudley-type regret bound similar to the one obtained in a non-constructive fashion by \cite{RaSr-14-OnlineNonparametricRegression}. Our regret bound is expressed in terms of the metric entropy in the sup norm, which yields optimal guarantees when the metric and sequential entropies are of the same order of magnitude. In particular our algorithm is the first one that achieves optimal rates for online regression over H\"{o}lder balls. In addition we show for this example how to adapt our chaining algorithm to get a reasonable computational efficiency with similar regret guarantees (up to a log factor).
\end{abstract}


%!TEX root = colt2015-onlinenonparametric.tex
\section{Introduction}

We consider the setting of online nonparametric regression for arbitrary deterministic sequences, which unfolds as follows. First, the environment chooses a sequence of observations
$(y_t)_{t \geqslant 1}$ in $\R$ and a sequence of input vectors
$(x_t)_{t \geqslant 1}$ in $\cX$, both initially hidden from the
forecaster.  At each time instant $t \in \mathbb{N}^* =
\{1,2,\ldots\}$, the environment reveals the data $x_t \in
\cX$; the forecaster then gives a prediction $\hat y_t \in \R$;
the environment in turn reveals the observation $y_t \in \R$; and finally, the forecaster incurs the square loss $(y_t - \hat{y}_t)^2$.

The term online \emph{nonparametric} regression means that we are interested in forecasters whose regret\\[-0.7cm]
\[
\Reg_T(\cF) \eqdef \sum_{t=1}^T \bigl(y_t - \hat{y}_t\bigr)^2 - \inf_{f \in \cF} \sum_{t=1}^T \bigl(y_t - f(x_t)\bigr)^2
\]
over standard nonparametric function classes $\cF \subseteq \R^{\cX}$ is as small as possible. In this paper we design and study an algorithm that achieves a regret bound of the form
\begin{equation}
\Reg_T(\cF) \leq c_1 B^2 \bigl(1+\log \cN_{\infty}(\cF,\gamma)\bigr) + c_2 B \sqrt{T} \int_0^{\gamma} \sqrt{\log \cN_{\infty}(\cF,\epsilon)} d \epsilon~,
\label{eq:DudleyRegretBound}
\end{equation}
where $\gamma \in \bigl(\frac{B}{T},B\bigr)$ is a parameter of the algorithm, where $B$ is an upper bound on $\max_{1 \leq t \leq T} |y_t|$, and where $\log \cN_{\infty}(\cF,\epsilon)$ denotes the metric entropy of the function set $\cF$ in the sup norm at scale~$\epsilon$ (cf.\ Section~\ref{sec:definitions}). %The above regret bound will be proved under the entropy growth assumption $\log \cN_{\infty}(\cF,\epsilon) \leq \epsilon^{-p}$ for some $p \in (0,2)$. The numerical constants $c_1$ and $c_2$ are made explicit in the proofs.

The integral on the right-hand side of \eqref{eq:DudleyRegretBound} is very close to what is known in probability theory as \emph{Dudley's entropy integral}, a useful tool to upper bound the expectation of a centered stochastic process with subgaussian increments (see, e.g., \citealt{Tal-05-GenericChaining,BoLuMa-12-Concentration}). In statistical learning (with i.i.d.\ data), Dudley's entropy integral is key to derive risk bounds on empirical risk minimizers; see, e.g., \cite{Massart03StFlour,RaSrTs-13-MinimaxRegretRisk}.

Very recently \cite{RaSr-14-OnlineNonparametricRegression} showed that the same type of entropy integral appears naturally in regret bounds for online nonparametric regression. The most part of their analysis is non-constructive in the sense that their regret bounds are obtained without explicitly constructing an algorithm. (Though they provide an abstract relaxation recipe for algorithmic purposes, we were not able to turn it into an explicit algorithm for online regression over nonparametric classes such as Hölder balls.)

One of our main contributions is to provide an explicit algorithm that achieves the regret bound~\eqref{eq:DudleyRegretBound}.
%{\color{red} Though the relaxation recipe of \cite{RaSr-14-OnlineNonparametricRegression} also satisfies \eqref{eq:DudleyRegretBound}, we were not able to use it for online nonparametric regression over Hölder balls.}
We note however that our regret bounds are in terms of a weaker notion of entropy, namely, metric entropy instead of the smaller (and optimal) sequential entropy. Fortunately, both notions are of the same order of magnitude for a reasonable number of examples, such as the ones outlined just below. We leave the question of modifying our algorithm to get sequential entropy regret bounds for future work.

The regret bound \eqref{eq:DudleyRegretBound}---that we call \emph{Dudley-type regret bound} thereafter---can be used to obtain optimal regret bounds for several classical nonparametric function classes. Indeed, when $\cF$ has a metric entropy $\log \cN_{\infty}(\cF,\epsilon) \leq C_p\epsilon^{-p}$ with\footnote{\label{foo:pgeq2}When $p>2$, we can also derive Dudley-type regret bounds that lead to a regret of $\mathcal{O}\bigl(T^{1-1/p}\bigr)$ in the same spirit as in \citet{RaSr-14-OnlineNonparametricRegression}. We omitted this case to ease the presentation.} $p \in (0,2)$, the bound \eqref{eq:DudleyRegretBound} entails
\begin{align}
\Reg_T(\cF) & \leq c_1 B^2 + c_1 B^2 C_p \gamma^{-p} + c_2 B \sqrt{C_p T} \int_0^{\gamma} \epsilon^{-p/2} d \epsilon \nonumber \\
& = c_1 B^2 + c_1 B^2 C_p \gamma^{-p} + \frac{2 \, c_2 B}{2-p} \sqrt{C_p T} \, \gamma^{1-p/2} = \mathcal{O}\bigl(T^{p/(p+2)}\bigr) \label{eq:DudleyRegretBound-csq}
\end{align}
for the choice of $\gamma = \Theta\bigl(T^{-1/(p+2)}\bigr)$. An example is given by  H\"{o}lder classes $\cF$ with regularity $\beta>1/2$ (cf. \citealt[Def~1.2]{Tsy-09-NonParametric}). We know from \citep{KolmogorovTikhomirov1961} or~\citep[Theorem~2]{Lor-62-MetricEntropy} that they satisfy $\log \cN_{\infty}(\cF,\epsilon) = \mathcal{O}\bigl(\epsilon^{-1/\beta}\bigr)$. Therefore, \eqref{eq:DudleyRegretBound-csq} entails a regret bound $\Reg_T(\cF) = \mathcal{O}\bigl(T^{1/(2\beta+1)}\bigr)$, which is in a way optimal since it corresponds to the optimal (minimax) quadratic risk $T^{-2\beta/(2\beta+1)}$ in statistical estimation with i.i.d.\ data.
%In Section~\ref{sec:Holder} we will come back to this particular case and prove a quasi-optimal regret of $T^{1/3} \log T$ for Lipschtiz classes, but with a manageable computational complexity.

\subsection{Why a simple Exponentially Weighted Average forecaster is not sufficient}

A natural approach (see \citealt{Vov-06-MetricEntropyOnlinePrediction}) to compete against a nonparametric class $\cF$ relies in running an Exponentially Weighted Average forecaster (EWA, see \citealt[p.14]{cesa-bianchi06prediction}) on an $\epsilon$-net $\cF^{(\epsilon)}$ of $\cF$ of finite size $\cN_{\infty}(\cF, \epsilon)$. This yields a regret bound of order $\epsilon T + \log \cN_{\infty}(\cF,\epsilon)$. The first term $\epsilon T$ is due to the approximation of $\cF$ by $\cF^{(\epsilon)}$, while the second term is the regret suffered by EWA on the finite class of experts $\cF^{(\epsilon)}$. As noted by \citet[Remark 11]{RaSr-14-OnlineNonparametricRegression}, the above regret bound is suboptimal for large nonparametric classes $\cF$. Indeed, for a metric entropy of order $\epsilon^{-p}$ with $p \in (0,2)$, optimizing the above regret bound in $\epsilon$ entails a regret of order $\mathcal{O}(T^{p/(p+1)})$ when \eqref{eq:DudleyRegretBound} yields the better rate $\mathcal{O}(T^{p/(p+2)})$.

\subsection{Constructing an online algorithm via the chaining technique}

Next we explain how the chaining technique from \citet{Dudley1967} (see appendix~\ref{sec:chaining-reminder} for a brief reminder) can be used to build an algorithm that satisfies a Dudley-type regret bound~\eqref{eq:DudleyRegretBound}. We approximate any function $f \in \cF$ by a sequence of refining approximations $\pi_0(f) \in \cF^{(0)},\pi_1(f) \in \cF^{(1)},\dots$, such that for all $k\geq 0$, $\sup_f \|\pi_k(f) - f\|_{\infty} \leq \gamma /2^{k}$ and $\card \cF^{(k)} = \cN_{\infty}(\cF, \gamma/2^k)$, so that:\\[-0.3cm]
\[
	  \inf_{f \in \cF} \sum_{t=1}^T \bigl(y_t - f(x_t)\bigr)^2 =  \inf_{f \in \cF} \sum_{t=1}^T \biggl(y_t - \pi_0(f)(x_t) - \sum_{k=0}^{\infty} \underbrace{\big[\pi_{k+1}(f) - \pi_k(f) \big](x_t) \bigg.}_{\hspace*{50pt} \norm{\cdot}_\infty \leq 3\gamma/2^{k+1}}\biggr)^2 \,.
\]
We use the above decomposition in Algorithm~\ref{alg:chainingEWA} (Section~\ref{sec:chainingEWA}) by performing two simultaneous aggregation tasks at two different scales:
\begin{itemize}[topsep=.1cm,parsep=.05cm,itemsep=.05cm]
    \item high-scale aggregation: we run an Exponentially Weighted Average forecaster to be competitive against every function $\pi_0(f)$ in the coarsest set $\cF^{(0)}$;
    \item low-scale aggregation: we run in parallel many instances of (an extension of) the Exponentiated Gradient (EG) algorithm so as to be competitive against the increments $\pi_{k+1}(f) - \pi_k(f)$. The advantage of using EG
%an Exponentiated Gradient algorithm
is that even if the number $N^{(k)}$ of increments $\pi_{k+1}(f) - \pi_k(f)$ is large for small scales $\epsilon$, the size of the gradients is very small, hence a manageable regret.
\end{itemize}

At the core of the algorithm lies the Multi-variable Exponentiated Gradient algorithm (Algorithm~\ref{alg:MultivarEG}) that makes it possible to perform low-scale aggregation at all scales $\epsilon<\gamma$ simultaneously.

\subsection{Comparison to previous works and main contributions}

\paragraph{Earlier uses of chaining and related techniques} Several ideas that we use in this paper were already present in the literature. \citet{OpHa-97-WorstCasePredictionLogLoss} and \citet{CBLu-01-LogLossMetricEntropy} derived Dudley-type regret bounds for the log loss using a two-scale aggregation and chaining arguments. At small scales, their algorithm is very specific to the log loss and it is unclear how to extend it to other exp-concave loss functions such as the square loss. Besides, they only use the chaining technique in their analysis by reducing the regret to an expected supremum, in the same spirit as \citet{RaSrTs-13-MinimaxRegretRisk} (square loss, batch setting) and \citet{RaSr-14-OnlineNonparametricRegression} (square loss, online learning with individual sequences). On the contrary, \citet{CeLu-99-PredictionIndiv} built an algorithm via chaining ideas (they use discretization sets $\cF^{(k)}$ similar to those above). However, their algorithm is specific to linear loss functions (e.g., absolute loss with binary observations), so that no linearization step and no high-scale aggregation are required.

\paragraph{Other papers on online learning with nonparametric classes} Related works also include the paper by \citet{Vov-06-MetricEntropyOnlinePrediction} where---for the problem under consideration here---suboptimal regret bounds are derived with the Exponentially Weighted Average forecaster. Another example of paper that addressed online learning over nonparametric function classes is the one by \citet{HaMe-07colt-OnlineLearningPriorKnowledge}. They also studied the regret with respect to the set of Lipschitz functions on $[0,1]^d$, but their loss functions are Lipschitz, hence their slower rates compared to ours.

\paragraph{Main contributions and outline of the paper} Our contributions are threefold: we first design the Multi-variable Exponentiated Gradient algorithm (Section~\ref{sec:MultivarEG}) which is crucial for the linearization step at all small scales simultaneously. We then present our main algorithm and derive a Dudley-type regret bound as in~\eqref{eq:DudleyRegretBound} (Section~\ref{sec:chainingEWA}). This general algorithm is computationally intractable for nonparametric classes. In Section~\ref{sec:Holder} we design an efficient algorithm in the case of H\"{o}lder classes. To the best of our knowledge, this is the first time the chaining technique has been used in a concrete fashion for individual sequences. Some proofs are postponed to the appendix.

\subsection{Some useful definitions}
\label{sec:definitions}

Let $\cF \subseteq \R^{\cX}$ be a set of bounded functions endowed with the sup norm $\Vert f \Vert_{\infty} \eqdef \sup_{x \in \cX} |f(x)|$. For all $\epsilon>0$, we call \emph{proper $\epsilon$-net} any subset $\cG \subseteq \cF$ such that $ \forall f \in \cF, \; \exists g \in \cG: \; \Vert f-g \Vert_{\infty} \leq \epsilon$. (If $\cG \not \subseteq \cF$, we call it \emph{non-proper}.) The cardinality of the smallest proper $\epsilon$-net is denoted by $\cN_{\infty}(\cF,\epsilon)$, and the logarithm $\log \cN_{\infty}(\cF,\epsilon)$ is called the \emph{metric entropy of $\cF$ at scale $\epsilon$.} When this quantity is finite for all $\epsilon>0$, we say that $\bigl(\cF,\Vert \cdot \Vert_{\infty}\bigr)$ is \emph{totally bounded}.

%!TEX root = colt2015-onlinenonparametric.tex
%\vspace{-0.2cm}
\section{The Chaining Exponentially Weighted Average Forecaster}

In this section we design an online algorithm---the \emph{Chaining Exponentially Weighted Average forecaster}---that achieves the Dudley-type regret bound~\eqref{eq:DudleyRegretBound}.
%It uses ideas borrowed from the chaining technique that we recalled in the introductory paragraphs.
In Section~\ref{sec:MultivarEG} below, we first define a subroutine that will prove crucial in our analysis, and \mbox{whose applicability may extend beyond this paper.}

%In Section~\ref{sec:MultivarEG} below, we first describe a new algorithm that will serve as a subroutine for the Chaining Exponentially Weighted Average forecaster, whose definition appears in Section~\ref{sec:chainingEWA}.

\subsection{Preliminary: the Multi-variable Exponentiated Gradient Algorithm}
\label{sec:MultivarEG}

Let $\Delta_{N} \eqdef \left\{\bu \in\R_+^{N}: \, \sum_{i=1}^{N} u_i = 1 \right\} \subseteq \R^{N}$ denote the simplex in $\R^N$. In this subsection we define and study a new extension of the Exponentiated Gradient algorithm \citep{KiWa97EGvsGD,CB99AnalysisGradientBased}. This extension is meant to minimize a sequence of multi-variable loss functions $\bigl(\bu^{(1)},\ldots,\bu^{(K)}\bigr) \mapsto \ell_t\bigl(\bu^{(1)},\ldots,\bu^{(K)}\bigr)$ simultaneously over all the variables $(\bu^{(1)},\ldots,\bu^{(K)}) \in \Delta_{N_1} \times \ldots \times \Delta_{N_K}$.

Our algorithm is described as Algorithm~\ref{alg:MultivarEG} below. We call it \emph{Multi-variable Exponentiated Gradient}. When $K=1$, it boils down to the classical Exponentiated Gradient algorithm over the simplex $\Delta_{N_1}$. But when $K \geq 2$, it performs $K$ simultaneous optimization updates (one for each direction $\bu^{(k)}$) that lead to a global optimum by joint convexity of the loss functions $\ell_t$.

%Note also that to simplify the presentation, we only use constant tuning parameters $\eta^{(k)}$

\begin{algorithm2e}[ht]
\SetKwInOut{Input}{input}
\SetKwInOut{Init}{initialization}

\Input{optimization domain $\Delta_{N_1}\times \ldots \times \Delta_{N_K}$ and tuning parameters $\eta^{(1)},\ldots,\eta^{(K)} >0$.}

\vspace{0.2cm}

\Init{set $\hat{\bu}^{(k)}_1 \eqdef \bigl(\frac{1}{N_k},\ldots,\frac{1}{N_k}\bigr) \in \Delta_{N_k}$ for all $k=1,\ldots,K$.}

\vspace{0.2cm}

\For{each round $t=1,2,\ldots$}{
\begin{itemize}
	\item Output $\bigl(\hat{\bu}^{(1)}_t,\ldots,\hat{\bu}^{(K)}_t\bigr) \in \Delta_{N_1}\times \ldots \times \Delta_{N_K}$ and observe the differentiable and jointly convex loss function $\ell_t:\Delta_{N_1}\times \ldots \times \Delta_{N_K} \to \R$.
	\item Compute the new weight vectors $\bigl(\hat{\bu}^{(1)}_{t+1},\ldots,\hat{\bu}^{(K)}_{t+1}\bigr) \in \Delta_{N_1}\times \ldots \times \Delta_{N_K}$ as follows:\\[-0.2cm]
	\[
	\hat{\bu}^{(k)}_{t+1,i} \eqdef \frac{\exp\!\left(-\eta^{(k)} \displaystyle \sum_{s=1}^t \partial_{\hat{u}^{(k)}_{s,i}} \ell_s\!\left(\hat{\bu}^{(1)}_s,\ldots,\hat{\bu}^{(K)}_s\right)\right)}{Z^{(k)}_{t+1}}~, \quad i \in \{1,\ldots,N_k\},
	\]
	where $\partial_{\hat{u}^{(k)}_{s,i}} \ell_s$ is the partial derivative of $\ell_s$ with respect to the $i$-th component of~$\hat{\bu}^{(k)}_{s}$, and where the normalizing factor is $Z^{(k)}_{t+1} \eqdef \sum_{i=1}^{N_k} \exp\!\left(-\eta^{(k)} \sum_{s=1}^t \partial_{\hat{u}^{(k)}_{s,i}} \ell_s\!\left(\hat{\bu}^{(1)}_s,\ldots,\hat{\bu}^{(K)}_s\right)\right)$.
%the normalization factor $Z^{(k)}_{t+1}$ is defined by
	%\[
	%Z^{(k)}_{t+1} \eqdef \sum_{i=1}^{N_k} \exp\!\left(-\eta^{(k)} \sum_{s=1}^t \partial_{\hat{u}^{(k)}_{s,i}} \ell_s\!\left(\hat{\bu}^{(1)}_s,\ldots,\hat{\bu}^{(K)}_s\right)\right)~.
	%\]
	\vspace{-0.8cm}
\end{itemize}
}
\caption{\label{alg:MultivarEG}Multi-variable Exponentiated Gradient}
\end{algorithm2e}

The Multi-variable Exponentiated Gradient algorithm satisfies the regret bound of Theorem~\ref{thm:MultivarEG} below. We first need some notations. We define the \emph{partial gradients}\\[-0.2cm]
\[
\nabla_{\bu^{(k)}} \ell_t = \Bigl( \partial_{u^{(k)}_1} \ell_t, \ldots, \partial_{u^{(k)}_{N_k}} \ell_t\Bigr)~, \qquad 1 \leq k \leq K~,
\]
\ \\[-0.3cm]
where $\partial_{u^{(k)}_i} \ell_t$ denotes the partial derivative of $\ell_t$ with respect to the scalar variable $u^{(k)}_i$. %$i \in \{1,\ldots,N_k\}$.
%Thus $\nabla_{\bu^{(k)}} \ell_t$ can be seen as the partial gradient of $\ell_t$ when $\ell_t$ is seen as a function of $\bu^{(k)} = \left(u^{(k)}_1,\ldots, u^{(k)}_{N_k}\right) \in \Delta_{N_k}$ only  (all the other variables $\bu^{(j)}$ for $j \neq k$ being kept equal).
Note that $\nabla_{\bu^{(k)}} \ell_t$ is a function that maps $\Delta_{N_1}\times \ldots \times \Delta_{N_K}$ to $\R^{N_k}$. Next we also use the notation\\[-0.2cm]
\[
\norm{\phi}_{\infty} \eqdef \sup_{\bu^{(1)},\ldots,\bu^{(K)}} \max_{1 \leq i \leq N_k} \Big|\phi_i\bigl(\bu^{(1)},\ldots,\bu^{(K)}\bigr)\Big|
\]
\ \\[-0.4cm]
for the sup norm of any vector-valued function $\phi:\Delta_{N_1}\times \ldots \times \Delta_{N_K} \to \R^{N_k}$, $1 \leq k \leq K$.

\begin{theorem}
\label{thm:MultivarEG}
Assume that the loss functions $\ell_t:\Delta_{N_1}\times \ldots \times \Delta_{N_K} \to \R$, $t \geq 1$, are differentiable and jointly convex. Assume also the following upper bound on their partial gradients: for all $k \in \{1,\ldots,K\}$, \\[-.7cm]
\begin{equation}
\max_{1 \leq t \leq T} \norm{\nabla_{\bu^{(k)}} \ell_t}_{\infty} \leq G^{(k)}~.
\label{eq:MultivarEG-regret-assumptiongradients}
\end{equation}
Then, the Multi-variable Exponentiated Gradient algorithm (Algorithm~\ref{alg:MultivarEG}) tuned with the parameters $\eta^{(k)} = \sqrt{2 \log(N_k)/T} \, / G^{(k)}$ has a regret bounded as follows:\\[-.7cm]
\begin{align*}
\sum_{t=1}^T \ell_t\!\left(\hat{\bu}^{(1)}_t,\ldots,\hat{\bu}^{(K)}_t\right) - \min_{\bu^{(1)},\ldots,\bu^{(K)}} \sum_{t=1}^T \ell_t\!\left(\bu^{(1)},\ldots,\bu^{(K)}\right) \leq \sqrt{2 T} \, \sum_{k=1}^K G^{(k)} \sqrt{\log N_k}~,
%\sum_{t=1}^T \ell_t\!\left(\hat{\bu}^{(1)}_t,\ldots,\hat{\bu}^{(K)}_t\right) - \min_{\bu^{(1)},\ldots,\bu^{(K)}} \sum_{t=1}^T \ell_t\!\left(\bu^{(1)},\ldots,\bu^{(K)}\right) \leq \sqrt{2} \sum_{k=1}^K \sqrt{\sum_{t=1}^T \norm{\nabla_{\bu^{(k)}} \ell_t}_{\infty}^2 \log N_k}~,
\end{align*}
where the minimum is taken over all $\left(\bu^{(1)},\ldots,\bu^{(K)}\right) \in \Delta_{N_1}\times \ldots \times \Delta_{N_K}$.
\end{theorem}
The proof of Theorem~\ref{thm:MultivarEG} is postponed to Appendix~\ref{sec:proofs-multivarEG}.

\subsection{The Chaining Exponentially Weighted Average Forecaster}
\label{sec:chainingEWA}

In this section we introduce our main algorithm: the \emph{Chaining Exponentially Weighted Average forecaster}. A precise definition will be given in Algorithm~\ref{alg:chainingEWA} below. For the sake of clarity, we first describe the main ideas underlying this algorithm.

Recall that we aim at proving a regret bound of the form~\eqref{eq:DudleyRegretBound}, whose right-hand side consists of two main terms:\\[-0.4cm]
\[
B^2 \log \cN_{\infty}(\cF,\gamma) \qquad \textrm{and} \qquad  B \sqrt{T} \int_0^{\gamma} \sqrt{\log \cN_{\infty}(\cF,\epsilon)} d \epsilon~.
\]
Our algorithm performs aggregation at two different levels: one level (at all scales $\epsilon \in (0, \gamma]$) to get the entropy integral above, and another level (at scale $\gamma$) to get the other term $B^2 \log \cN_{\infty}(\cF,\gamma)$. More precisely:
\begin{itemize}[topsep=.1cm,itemsep=.05cm,parsep=.05cm]
	\item for all $k \in \N$, let $\cF^{(k)}$ be a proper $\gamma/2^k$-net of $(\cF,\norm{\cdot}_{\infty})$ of minimal cardinality\footnote{We assume that $(\cF,\norm{\cdot}_{\infty})$ is totally bounded.} $\cN_{\infty}\bigl(\cF,\gamma/2^k\bigr)$;
	\item for all $k \geq 1$, set $\cG^{(k)} \eqdef \{\pi_k(f)-\pi_{k-1}(f): \; f \in \cF\}$, where
	\[
	\forall f \in \cF, \quad \pi_k(f) \in \argmin_{h \in \cF^{(k)}} \norm{f-h}_{\infty}~.
	\]
\end{itemize}

\ \\[-0.7cm]
We denote:
\begin{itemize}[topsep=.1cm,itemsep=.1cm,parsep=.1cm]
	\item the elements of $\cF^{(0)}$ by $f^{(0)}_1,\ldots,f^{(0)}_{N_0}$ with $N_0 = \cN_{\infty}\bigl(\cF,\gamma\bigr)$;
	\item the elements of $\cG^{(k)}$ by $g^{(k)}_1,\ldots,g^{(k)}_{N_k}$; note that $N_k \leq \cN_{\infty}\bigl(\cF,\gamma/2^k\bigr) \cN_{\infty}\bigl(\cF,\gamma/2^{k-1}\bigr)$.
	\end{itemize}

\medskip \noindent
With the above definitions, our algorithm can be described as follows:
\begin{enumerate}[topsep=.1cm]
	\item Low-scale aggregation: for every $j \in \{1,\ldots,N_0\}$, we use a Multi-variable Exponentiated Gradient forecaster to mimic the best predictor in the neighborhood of $f^{(0)}_j$: we set, at each round $t \geq 1$, \\[-.5cm]
	\begin{equation}
		\hat{f}_{t,j} \eqdef f^{(0)}_j + \sum_{k=1}^K \sum_{i=1}^{N_k} \hat{u}_{t,i}^{(j,k)} g^{(k)}_i~,
		\label{eq:chainingEWA-intermediatepredictors}
	\end{equation}
	where $K \eqdef \bigl\lceil \log_2(\gamma T/B) \bigr\rceil$, so that the lowest scale is $\gamma/2^K \approx B/T$. The above weight vectors $\hat{u}_{t}^{(j,k)} \in \Delta_{N_k}$ are defined in Equation~\eqref{eq:chainingEWA-weights-lowscale} of Algorithm~\ref{alg:chainingEWA}. They correspond exactly to the weight vectors output by the Multi-variable Exponentiated Gradient forecaster (Algorithm~\ref{alg:MultivarEG}) applied to the loss functions $\ell^{(j)}_t:\Delta_{N_1} \times \ldots \times \Delta_{N_K} \to \R$ defined for all $t \geq 1$ ($j$ is fixed)~by
	\begin{align}
	\ell^{(j)}_t\!\left(\bu^{(1)},\ldots,\bu^{(K)}\right) & = \left(y_t - f^{(0)}_j(x_t) - \sum_{k=1}^K \sum_{i=1}^{N_k} u_{i}^{(k)} g^{(k)}_i(x_t) \right)^2~. \label{eq:chainingEWA-applicationMEG}	%\ell^{(j)}_t\!\left(\bu^{(j,1)}_t,\ldots,\bu^{(j,K)}_t\right) & = \left(y_t - \hat{f}_{j,t}(x_t)\right)^2 = \left(y_t - f^{(0)}_j(x_t) - \sum_{k=1}^K \sum_{i=1}^{N_k} \hat{u}_{t,i}^{(j,k)} g^{(k)}_i(x_t) \right)^2~. \label{eq:chainingEWA-applicationMEG}
	\end{align}
	\item High-scale aggregation: we use a standard Exponentially Weighted Average forecaster to aggregate all the $\hat{f}_{t,j}$, $j=1,\ldots,N_0$, as follows:\\[-0.7cm]
	\[ 
	 \hspace*{5cm} \hat{f}_t = \sum_{j=1}^{N_0} \hat{w}_{t,j} \hat{f}_{t,j}~,
	\]
	%\ \\[-0.2cm]
	where the weights $\hat{w}_{t,j}$ are defined in Equation~\eqref{eq:chainingEWA-weights-highscale} of Algorithm~\ref{alg:chainingEWA}. At time $t$, our algorithm predicts $y_t$ with $\hat{y}_t \eqdef \hat{f}_t(x_t)$.
\end{enumerate}

\begin{algorithm2e}[t!]
\SetKwInOut{Input}{input}
\SetKwInOut{Init}{initialization}

\Input{maximal range $B>0$, tuning parameters $\eta^{(0)},\eta^{(1)},\ldots,\eta^{(K)} >0$,\\
high-scale functions $f^{(0)}_j: \cX \to \R$ for $1 \leq j \leq N_0$, \\
low-scale functions $g^{(k)}_i: \cX \to \R$ for $k \in \{1,\ldots,K\}$ and $i \in \{1,\ldots,N_k\}$.}

\vspace{0.2cm}

\Init{set $\hat{\bw}_1 = \bigl(\frac{1}{N_0},\ldots,\frac{1}{N_0}\bigr) \in \Delta_{N_0}$ and\\
$\hat{\bu}^{(j,k)}_1 \eqdef \bigl(\frac{1}{N_k},\ldots,\frac{1}{N_k}\bigr) \in \Delta_{N_k}$ for all $j \in \{1,\ldots,N_0\}$ and $k \in \{1,\ldots,K\}$.}

\vspace{0.2cm}

\For{each round $t=1,2,\ldots$}{
\vspace{0.1cm}
\begin{itemize}[topsep=.2cm,itemsep=.1cm,parsep=.1cm]
	\item Define the aggregated functions $\hat{f}_{t,j}: \cX \to \R$ for all $j \in \{1,\ldots,N_0\}$ by\\[-0.3cm]
	\[
	\hat{f}_{t,j} \eqdef  f^{(0)}_j + \sum_{k=1}^K \sum_{i=1}^{N_k} \hat{u}^{(j,k)}_{t,i} g^{(k)}_i~.
	\]
	\vspace{-0.5cm}
	\item Observe $x_t \in \cX$, predict $\displaystyle \hat{y}_t = \sum_{j=1}^{N_0} \hat{w}_{t,j} \hat{f}_{t,j}(x_t)$, and observe $y_t \in [-B,B]$.
	%\item Observe $x_t \in \cX$, predict
	%\[
	%\hat{y}_t = \sum_{j=1}^{N_0} \hat{w}_{t,j} \hat{f}_{t,j}(x_t)~,
	%\]
	%and observe $y_t \in [-B,B]$.
	\item Low-scale update: compute the new weight vectors $\hat{\bu}^{(j,k)}_{t+1} = \Bigl(\hat{u}^{(j,k)}_{t+1,i}\Bigr)_{1 \leq i \leq N_k} \in \Delta_{N_k}$ for all $j \in \{1,\ldots,N_0\}$ and $k \in \{1,\ldots,K\}$ as follows:\\[-0.5cm]
	%\[
	%\hat{\bu}^{(j,k)}_{t+1,i} \eqdef \frac{1}{Z^{(j,k)}_{t+1}} \exp\!\left(-\eta^{(k)} \displaystyle \sum_{s=1}^t -2\Bigl(y_s-\hat{f}_{s,j}(x_s)\Bigr) g^{(k)}_i(x_s) \right)~, \quad i \in \{1,\ldots,N_k\},
	%\]
	%where the normalizing constant $Z^{(j,k)}_{t+1}$ is defined by
	%\[
	%Z^{(j,k)}_{t+1} \eqdef \sum_{i=1}^{N_k} \exp\!\left(-\eta^{(k)} \displaystyle \sum_{s=1}^t -2\Bigl(y_s-\hat{f}_{s,j}(x_s)\Bigr) g^{(k)}_i(x_s) \right)~.
	%\]
	%
	\begin{equation}
	\label{eq:chainingEWA-weights-lowscale}
	\hat{u}^{(j,k)}_{t+1,i} \eqdef \frac{\exp\!\left(-\eta^{(k)} \displaystyle \sum_{s=1}^t -2\Bigl(y_s-\hat{f}_{s,j}(x_s)\Bigr) g^{(k)}_i(x_s) \right)}{\displaystyle \sum_{i'=1}^{N_k} \exp\!\left(-\eta^{(k)} \displaystyle \sum_{s=1}^t -2\Bigl(y_s-\hat{f}_{s,j}(x_s)\Bigr) g^{(k)}_{i'}(x_s) \right)}~, \quad i \in \{1,\ldots,N_k\}~.
	\end{equation}
	\item High-scale update: compute the new weight vector $\hat{\bw}_{t+1} = \bigl(\hat{w}_{t+1,j}\bigr)_{1 \leq j \leq N_0} \in \Delta_{N_0}$ as follows:\\[-0.7cm]
	\begin{equation}
	\label{eq:chainingEWA-weights-highscale}
	\hat{w}_{t+1,j} \eqdef \frac{\exp\!\left(-\eta^{(0)} \displaystyle \sum_{s=1}^t \Bigl(y_s-\hat{f}_{s,j}(x_s)\Bigr)^2 \right)}{\displaystyle \sum_{j'=1}^{N_0} \exp\!\left(-\eta^{(0)} \sum_{s=1}^t \Bigl(y_s-\hat{f}_{s,j'}(x_s)\Bigr)^2 \right)}~, \quad j \in \{1,\ldots,N_0\}~.
	\end{equation}
	\vspace{-0.8cm}
	%where $[x]_B \eqdef \max\bigl\{ \min\{x,B\}, -B \bigr\}$ denotes the clipping of $x \in \R$ onto $[-B,B]$.
\end{itemize}
}
\caption{\label{alg:chainingEWA}Chaining Exponentially Weighted Average forecaster}
\end{algorithm2e}

\noindent
Next we show that the Chaining Exponentially Weighted Average forecaster satisfies a Dudley-type regret bound as in~\eqref{eq:DudleyRegretBound}.

\begin{theorem}
\label{thm:chainingEWA-regret}
Let $B>0$, $T \geq 1$, and $\gamma \in \bigl({B}/{T},B\bigr)$.
\begin{itemize}[topsep=.1cm,itemsep=.05cm,parsep=.05cm]
	\item Assume that $\max_{1 \leq t \leq T}|y_t| \leq B$ and that $\sup_{f \in \cF} \Vert f \Vert_{\infty} \leq B$.
	\item Assume that $(\cF,\norm{\cdot}_{\infty})$ is totally bounded and define $\cF^{(0)} = \bigl\{f^{(0)}_1, \ldots, f^{(0)}_{N_0} \bigr\}$ and $\cG^{(k)} = \bigl\{g^{(k)}_1, \ldots, g^{(k)}_{N_k} \bigr\}$, $k=1,\ldots,K$, as above.
\end{itemize} 

\smallskip \noindent
Then, the Chaining Exponentially Weighted Average forecaster (Algorithm~\ref{alg:chainingEWA}) tuned with the parameters $\eta^{(0)} = 1/(50 B^2)$ and $\eta^{(k)} = \sqrt{2 \log(N_k)/T} \, 2^k / (30 B \gamma)$ for all $k=1,\ldots,K$ satisfies:
\[
\Reg_T(\cF) \leq B^2 \bigl(5+50\log \cN_{\infty}(\cF,\gamma)\bigr) + 120 B \sqrt{T} \int_0^{\gamma/2} \sqrt{\log \cN_{\infty}(\cF,\epsilon)} d \epsilon~.
\]
\end{theorem}

As a corollary (cf.\ \eqref{eq:DudleyRegretBound-csq} in the introduction), when $\log \cN_{\infty}(\cF,\epsilon) \leq C_p\epsilon^{-p}$ with $p \in (0,2)$, the Chaining Exponentially Weighted Average forecaster tuned as above and with $\gamma = \Theta\bigl(T^{-1/(p+2)}\bigr)$ has a regret of $\mathcal{O}\bigl(T^{p/(p+2)}\bigr)$. { This in turn yields a regret of $\Reg_T(\cF) = \mathcal{O}\bigl(T^{1/(2\beta+1)}\bigr)$ when $\cF$ is the H\"{o}lder class with regularity $\beta>1/2$, which corresponds to the optimal (minimax) quadratic risk $T^{-2\beta/(2\beta+1)}$ in statistical estimation with i.i.d.\ data. We address the particular case of Hölder functions and the associated computational issues in Section~\ref{sec:Holder} and Appendix~\ref{sec:Holder-appendix} below.

Another corollary of Theorem~\ref{thm:chainingEWA-regret} can be drawn in the setting of sparse high-dimensional online linear regression, which is a particular case of a parametric class with $p \approx 0$. In the same spirit as in~\citet{Ger-11-SparsityRegretBounds} and in~\citet[Example~1]{RaSr-14-OnlineNonparametricRegression}, we consider $d$ features $\phi_1,\ldots,\phi_d:\cX \to [-B,B]$ and we define} 
$
 \cF = \big\{ {\textstyle \sum_{j=1}^d} u_j \phi_j: \; \bu \in \Delta_d, \norm{\bu}_0 = s \big\}
$
to be the set of all $s$-sparse convex combinations of the features ($\norm{\bu}_0$ denotes the number of non-zero coefficients of $\bu$). Then, using Theorem 1 of \citet{GaInYa-13-LqHulls} with $(M,q,p,r)=(s,1,+\infty,2)$ we can see that $\log \cN_{\infty}(\cF,\epsilon) \lesssim \log \binom{d}{s} + s \log\bigl(1+1/(\epsilon\sqrt{s})\bigr)$. Plugging this bound in Theorem~\ref{thm:chainingEWA-regret} with $\gamma = 1/\sqrt{T}$ yields a regret bound of order $s \log(1+dT/s)$. Thus, Theorem~\ref{thm:chainingEWA-regret} also yields (quasi) optimal rates for sparse high-dimensional online linear regression.

Finally, for much larger function classes with $p>2$, we could derive a slightly modified Dudley-type regret bound of $\mathcal{O}\bigl(T^{1-1/p}\bigr)$ with a slightly modified algorithm, in the same spirit as in \citet{RaSr-14-OnlineNonparametricRegression}. We omit this case for the sake of conciseness.

\begin{remark}
In Theorem~\ref{thm:chainingEWA-regret} above, we assumed that the observations $y_t$ and the predictions $f(x_t)$ are all bounded by $B$, and that $B$ is known in advance by the forecaster. We can actually remove this requirement by using adaptive techniques of \cite{GeYu-14-L1Balls}, namely, adaptive clipping of the intermediate predictions $\hat{f}_{t,j}(x_t)$ and adaptive Lipschitzification of the square loss functions $\ell^{(j)}_t$. This modification enables us to derive the same regret bound (up to multiplicative constant factors) with $B = \max_t |y_t|$, but without knowing $B$ in advance, and without requiring that $\sup_{f \in \cF} \Vert f \Vert_{\infty}$ is also upper bounded by $B$. Of course these adaptation techniques also make it possible to tune all parameters without knowing $T$ in advance.
\end{remark}

\begin{remark}
Even in the case when $B$ is known by the forecaster, the clipping and Lipschitzification techniques of \cite{GeYu-14-L1Balls} can be useful to get smaller constants in the regret bound. We could indeed replace the constants $50$ and $120$ with $8$ and $48$ respectively. (Moreover, the regret bound would also hold true for $\gamma > B$.) We chose however not to use these refinements in order to simplify the analysis.
\end{remark}

\begin{remark}
We assumed that the performance of a forecast $\hat y_t$ at round $t\geq 1$ is measured through the square loss $\ell_t(\hat y_t) = (\hat y_t - y_t)^2$, which is $1/(50B^2)$-exp-concave on $[-4B,4B]$. The analysis can easily be extended to all $\eta$-exp-concave (and thus convex) loss functions $\ell_t$ on $[-4B,4B]$ that also satisfy a self-bounding property of the form $\bigl|d\ell_t/d \hat{y}_t\bigr| \leq C \ell_t^r$ (an example is given by $\ell_t(\hat y_t) = \bigl|\hat y_t - y_t\bigr|^r$ with $r \geq 2$). The regret bound of Theorem~\ref{thm:chainingEWA-regret} remains unchanged up to a multiplicative factor depending on $B$, $C$, and $r$. If the loss functions $\ell_t$ are only convex (e.g., the absolute loss $\ell_t(\hat y_t) = |\hat y_t - y_t|$ or the pinball loss to perform quantile regression), the high-scale aggregation step is more costly: the term of order $\log \cN_\infty(\cF, \gamma)$ is replaced with a term of order $\sqrt{T \log \cN_\infty(\cF, \gamma)}$.
%it is not possible to perform a high scale aggregation whose cumulative regret can be bounded independently of $T$. The first main term of order $\log \cN_\infty(\cF, \gamma)$ is then replaced with a term of order $\sqrt{T \log \cN_\infty(\cF, \gamma)}$.
\end{remark}

\begin{proofref}{Theorem~\ref{thm:chainingEWA-regret}}
We split our proof into two parts---one for each aggregation level.\\[0.3cm]
\emph{Part 1: low-scale aggregation}. \\[0.2cm]
In this part, we fix $j \in \{1,\ldots,N_0\}$. As explained right before~\eqref{eq:chainingEWA-applicationMEG}, the tuple of weight vectors $\bigl(\hat{\bu}^{(j,1)}_{t},\ldots,\hat{\bu}^{(j,K)}_{t}\bigr)  \in \Delta_{N_1} \times \ldots \Delta_{N_K}$ computed at all rounds $t \geq 1$ corresponds exactly to the output of the Multi-variable Exponentiated Gradient forecaster when applied to the loss functions~$\ell^{(j)}_t$, $t \geq 1$, defined in~\eqref{eq:chainingEWA-applicationMEG}. We can therefore apply Theorem~\ref{thm:MultivarEG} after checking its assumptions:
\begin{itemize}[topsep=.1cm,itemsep=.1cm,parsep=.1cm]
	\item the loss functions~$\ell^{(j)}_t$ are indeed differentiable and jointly convex;
	\item the norms \smash{$\big\Vert\nabla_{\hat{\bu}^{(j,k)}_t} \ell^{(j)}_t\big\Vert_{\infty}$} of the partial gradients are bounded by $30 B \gamma/2^k$ for all $1~\leq~k~\leq~K$. Indeed, the $i$-th coordinate of $\nabla_{\hat{\bu}^{(j,k)}_t} \ell^{(j)}_t$ is equal to \\[-0.1cm]
	\begin{equation}
	\hspace*{3cm} \partial_{\hat{u}^{(j,k)}_{t,i}} \ell^{(j)}_t = -2\Bigl(y_t-\hat{f}_{t,j}(x_t)\Bigr) g^{(k)}_i(x_t),
	\label{eq:chainingEWA-regret-partialgradients}
	\end{equation}
	which can be upper bounded (in absolute value) by $2 \times 5B \times 3\gamma/2^k$. To see why this is true, first note that $\bigl|g^{(k)}_i(x_t)\bigr| \leq \big\Vert g^{(k)}_i \big\Vert_{\infty} = \big\Vert \pi_k(f) - \pi_{k-1}(f)\big\Vert_{\infty}$ for some $f \in \cF$ (by definition of $\cG^{(k)}$), so that, by the triangle inequality and by definition of $\pi_k(f)$ and $\cF^{(k)}$:
	\begin{equation}
	\left|g^{(k)}_i(x_t)\right| \leq \big\Vert \pi_k(f) - f\big\Vert_{\infty} + \big\Vert\pi_{k-1}(f) - f\big\Vert_{\infty} \leq \frac{\gamma}{2^k} + \frac{\gamma}{2^{k-1}} = \frac{3\gamma}{2^k}~.
	\label{eq:chainingEWA-regret-bound-gi}
	\end{equation}
	Second, note that $\bigl|y_t-\hat{f}_{t,j}(x_t)\bigr| \leq \bigl|y_t\bigr| + \bigl|\hat{f}_{t,j}(x_t)\bigr| \leq 5B$. Indeed, we have $|y_t|\leq B$ by assumption and, by definition of $\hat{f}_{t,j}$ in~\eqref{eq:chainingEWA-intermediatepredictors}, we also have
	\begin{align}
		\bigl|\hat{f}_{t,j}(x_t)\bigr| & \leq \norm{f^{(0)}_j}_{\infty} + \sum_{k=1}^K \sum_{i=1}^{N_k} \hat{u}_{t,i}^{(j,k)} \bigl|g^{(k)}_i(x_t)\bigr| \leq B + \sum_{k=1}^K \frac{3\gamma}{2^k} \leq B + 3\gamma \leq 4B~, \label{eq:chainingEWA-intermediatepredictors-norm}
	\end{align}
	where we used the inequalities $\big\Vert f^{(0)}_j \big\Vert_{\infty} \leq \sup_{f \in \cF} \Vert f \Vert_{\infty} \leq B$ (by assumption), and where we combined~\eqref{eq:chainingEWA-regret-bound-gi} with the fact that $\sum_{i=1}^{N_k} \hat{u}_{t,i}^{(j,k)}=1$. The last inequality above is obtained from the assumption $\gamma \leq B$. Substituting the above various upper bounds in~\eqref{eq:chainingEWA-regret-partialgradients} entails that $\big\Vert\nabla_{\hat{\bu}^{(j,k)}_t} \ell^{(j)}_t\big\Vert_{\infty} \leq 30 B \gamma/2^k$ for all $1 \leq k \leq K$, as claimed earlier.
\end{itemize}

\medskip \noindent
We are now in a position to apply Theorem~\ref{thm:MultivarEG}. It yields:\\[-.6cm]
\begin{align}
	\sum_{t=1}^T \left(y_t - \hat{f}_{t,j}(x_t)\right)^2 \leq & \inf_{g_1, \ldots, g_K} \sum_{t=1}^T \left(y_t - \left(f^{(0)}_j + g_1 + \ldots +g_K \right)(x_t)\right)^2 \nonumber \\
	& + \sqrt{2T} \sum_{k=1}^K 30 B \gamma/2^k \sqrt{\log N_k}~,
\label{eq:chainingEWA-regret-lowscalecontribution-1}
\end{align}
where the infimum is over all functions $g_1 \in \cG^{(1)}, \ldots, g_K \in \cG^{(K)}$ (we used the regret bound of Theorem~\ref{thm:MultivarEG} with Dirac weight vectors $\bu^{(k)} = \delta_{i_k}$, $i_k=1,\ldots,N_k$).

Now, using the fact that $N_k \leq \cN_{\infty}\bigl(\cF,\gamma/2^k\bigr) \cN_{\infty}\bigl(\cF,\gamma/2^{k-1}\bigr) \leq \bigl(\cN_{\infty}\bigl(\cF,\gamma/2^k\bigr)\bigr)^2$, we get
\begin{align*}
\sum_{k=1}^K \frac{\gamma}{2^{k}} \sqrt{\log N_k} & \leq 2\sqrt{2} \sum_{k=1}^K \left(\frac{\gamma}{2^{k}}-\frac{\gamma}{2^{k+1}}\right) \sqrt{\log \cN_{\infty}\bigl(\cF,\gamma/2^k\bigr)} \\
& \leq 2 \sqrt{2} \sum_{k=1}^K \int_{\gamma/2^{k+1}}^{\gamma/2^k} \sqrt{\log \cN_{\infty}\bigl(\cF,\epsilon\bigr)} d \epsilon \leq 2\sqrt{2} \int_0^{\gamma/2} \sqrt{\log \cN_{\infty}\bigl(\cF,\epsilon\bigr)} d \epsilon~,
\end{align*}
where the inequality before last follows by monotonicity of $\epsilon \mapsto \cN_{\infty}(\cF, \epsilon)$ on every interval $\bigl[\gamma/2^{k+1}, \gamma/2^k \bigr]$. Finally, substituting the above integral in~\eqref{eq:chainingEWA-regret-lowscalecontribution-1} yields
\begin{align}
	\sum_{t=1}^T \left(y_t - \hat{f}_{t,j}(x_t)\right)^2 \leq & \inf_{g_1, \ldots, g_K} \sum_{t=1}^T \left(y_t - \left(f^{(0)}_j + g_1 + \ldots +g_K \right)(x_t)\right)^2 \nonumber \\
	& + 120 B \sqrt{T} \int_0^{\gamma/2} \sqrt{\log \cN_{\infty}\bigl(\cF,\epsilon\bigr)} d \epsilon~.
\label{eq:chainingEWA-regret-lowscalecontribution-2}
\end{align}

\noindent
\emph{Part 2: high-scale aggregation}.\\[0.2cm]
The prediction $\hat{y}_t = \hat{f}_t(x_t) = \sum_{j=1}^{N_0} \hat{w}_{t,j} \hat{f}_{t,j}(x_t)$ at time $t$ is a convex combination of the intermediate predictions $\hat{f}_{t,j}(x_t)$, where the weights $\hat{w}_{t,j}$ correspond exactly to those of the standard Exponentially Weighted Average forecaster tuned with $\eta^{(0)} = 1/(50 B^2) = 1/\bigl(2(5B)^2\bigr)$. Since the intermediate predictions $\hat{f}_{t,j}(x_t)$ lie in $[-4B,4B]$ (by~\eqref{eq:chainingEWA-intermediatepredictors-norm} above), and since the square loss $z \mapsto (y_t-z)^2$ is $\eta^{(0)}$-exp-concave on $[-4B,4B]$ for any $y_t \in [-B,B]$, we get from Proposition~3.1 and Page~46 of \cite{cesa-bianchi06prediction} that \\[-.4cm]
\begin{align}
\sum_{t=1}^T \left(y_t - \hat{y}_t\right)^2 \leq & \min_{1 \leq j \leq N_0} \sum_{t=1}^T \left(y_t - \hat{f}_{t,j}(x_t)\right)^2 + \frac{\log N_0}{\eta^{(0)}} \nonumber \\
\leq & \inf_{f_0, g_1, \ldots, g_K} \sum_{t=1}^T \bigl(y_t - \left(f_0 + g_1 + \ldots +g_K \right)(x_t)\bigr)^2 \nonumber \\
	& + 120 B \sqrt{T} \int_0^{\gamma/2} \sqrt{\log \cN_{\infty}\bigl(\cF,\epsilon\bigr)} d \epsilon + 50 B^2 \log \cN_{\infty}\bigl(\cF,\gamma\bigr)~, \label{eq:chainingEWA-regret-highscale-1}
\end{align}
where the infimum is over all functions $f_0 \in \cF^{(0)}, g_1 \in \cG^{(1)}, \ldots, g_K \in \cG^{(K)}$. The last inequality above was a consequence of~\eqref{eq:chainingEWA-regret-lowscalecontribution-2}. Next we apply the chaining idea: by definition of the function sets $\cF^{(0)} \supseteq \{ \pi_0(f): \; f \in \cF\}$ and $\cG^{(k)} = \{ \pi_k(f)-\pi_{k-1}(f): \; f \in \cF\}$, we have \\[-1cm]

\begin{align}
& \inf_{f_0, g_1, \ldots, g_K} \sum_{t=1}^T \left(y_t - \bigl(f_0 + g_1 + \ldots +g_K \right)(x_t)\bigr)^2 \nonumber \\[-0.1cm]
& \qquad \leq \inf_{f \in \cF} \sum_{t=1}^T \Bigl(y_t - \bigl(\pi_0(f) + \bigl[\pi_1(f)-\pi_0(f)\bigr] + \ldots +\bigl[\pi_K(f)-\pi_{K-1}(f)\bigr] \bigr)(x_t)\Bigr)^2 \nonumber \\[-0.1cm]
& \qquad = \inf_{f \in \cF} \sum_{t=1}^T \bigl(y_t - \pi_K(f)(x_t)\bigr)^2 \nonumber \\[-0.1cm]
& \qquad \leq \inf_{f \in \cF} \sum_{t=1}^T \left[\bigl(y_t - f(x_t)\bigr)^2 + 2 \cdot 2B \norm{\pi_K(f)-f}_{\infty} + \norm{\pi_K(f)-f}_{\infty}^2 \right]\label{eq:chainingEWA-regret-highscale-2} \\[-0.1cm]
& \qquad \leq \inf_{f \in \cF} \sum_{t=1}^T \bigl(y_t - f(x_t)\bigr)^2 + 4B^2+\frac{B^2}{T}~, \label{eq:chainingEWA-regret-highscale-3}
\end{align}
\ \\[-0.3cm]
where \eqref{eq:chainingEWA-regret-highscale-2} is obtained by expanding the square $\bigl(y_t - \pi_K(f)(x_t)\bigr)^2 = \bigl(y_t - f(x_t) + f(x_t) - \pi_K(f)(x_t)\bigr)^2$, and where~\eqref{eq:chainingEWA-regret-highscale-3} follows from the fact that $\norm{\pi_K(f)-f}_{\infty} \leq \gamma/2^K \leq B/T$ by definition of $\pi_K(f)$ and $K=\bigl\lceil \log_2(\gamma T/B) \bigr\rceil$. Combining~\eqref{eq:chainingEWA-regret-highscale-1} and~\eqref{eq:chainingEWA-regret-highscale-3} concludes the proof.
\end{proofref}

\vspace{-0.2cm}

%!TEX root = colt2015-onlinenonparametric.tex
\vspace{-0.5cm}
\section{An efficient chaining algorithm for H\"{o}lder classes}
\label{sec:Holder}

The Chaining Exponentially Weighted Average forecaster of the previous section is quite natural since it explicitly exploits the $\epsilon$-nets that appear in the Dudley-type regret bound~\eqref{eq:DudleyRegretBound}. However its time and space computational complexities are prohibitively large (exponential in $T$) since it is necessary to update exponentially many weights at every round $t$. It actually turns out that, fortunately, most standard function classes have a sufficiently nice structure. This enables us to adapt the previous chaining technique on (quasi-optimal) $\epsilon$-nets that are much easier to exploit from an algorithmic viewpoint. We describe below the particular case of Lipschitz classes; the more general case of H\"{o}lder classes is postponed to Appendix~\ref{sec:Holder-appendix}.

In all the sequel, $\cF$ denotes the set of functions from $[0,1]$ to $[-B,B]$ that are $1$-Lipschitz. Recall from the introduction that $\log \cN_{\infty}(\cF,\epsilon) = \Theta(\epsilon^{-1})$, so that, by Theorem~\ref{thm:chainingEWA-regret} and~\eqref{eq:DudleyRegretBound-csq}, the Chaining Exponentially Weighted Average forecaster guarantees a regret of $\mathcal{O}\bigl(T^{1/3}\bigr)$. We explain below how to modify this algorithm with $\epsilon$-nets of $\bigl(\cF,\Vert \cdot \Vert_{\infty} \bigr)$ that are easier to manage from a computational viewpoint. This leads to a quasi-optimal regret of $\mathcal{O}\bigl(T^{1/3} \log T\bigr)$; see Theorem~\ref{thm:DyadicChaining}.

\subsection{Constructing computationally-manageable $\epsilon$-nets via a dyadic discretization}

Let $\gamma \in \bigl(\frac{B}{T},B\bigr)$ be a fixed real number that will play the same role as in Theorem~\ref{thm:chainingEWA-regret}. Using the fact that all functions in $\cF$ are $1$-Lipschitz, we can approximate $\cF$ with piecewise-constant functions as follows. We partition the $x$-axis $[0,1]$ into $1/\gamma$ subintervals $I_a \eqdef \bigl[(a-1) \gamma, a \gamma \bigr)$, $a=1,\ldots,1/\gamma$ (the last interval is closed at $x=1$). We also use a discretization of length $\gamma$ on the $y$-axis $[-B,B]$, by considering values of the form $c^{(0)} = -B+j \gamma$, $j=0,\ldots,2B/\gamma$. (For the sake of simplicity, we assume that both $1/\gamma$ and $2B/\gamma$ are integers.) We then define the set $\cF^{(0)}$ of piecewise-constant functions $f^{(0)}:[0,1] \to [-B,B]$ of the form\\[-0.3cm]
\begin{equation}
\label{eq:Lip-def-F0}
	\qquad f^{(0)}(x) = \sum_{a=1}^{1/\gamma} c_a^{(0)} \indicator{x \in I_a}~, \qquad c^{(0)}_1,\ldots,c^{(0)}_{1/\gamma} \in \cC^{(0)} \eqdef \left\{-B+j \gamma: j=0,\ldots,\frac{2B}{\gamma} \right\}~.
\end{equation}

Using the fact that all functions in $\cF$ are $1$-Lipschitz, it is quite straightforward to see that $\cF^{(0)}$ is a $\gamma$-net\footnote{This $\gamma$-net is not proper since $\cF^{(0)} \not\subseteq \cF$.} of $\bigl(\cF,\Vert \cdot \Vert_{\infty} \bigr)$. (To see why this is true, we can choose $c^{(0)}_a \in \argmin_{c \in C^{(0)}} \bigl|f(x_a)-c\bigr|$, where $x_a$ is the center of the subinterval~$I_a$. See Lemma~\ref{lem:Lip-gammanets} in the appendix for further details.)

\begin{figure}[ht]
\centering
\includegraphics[scale=0.75]{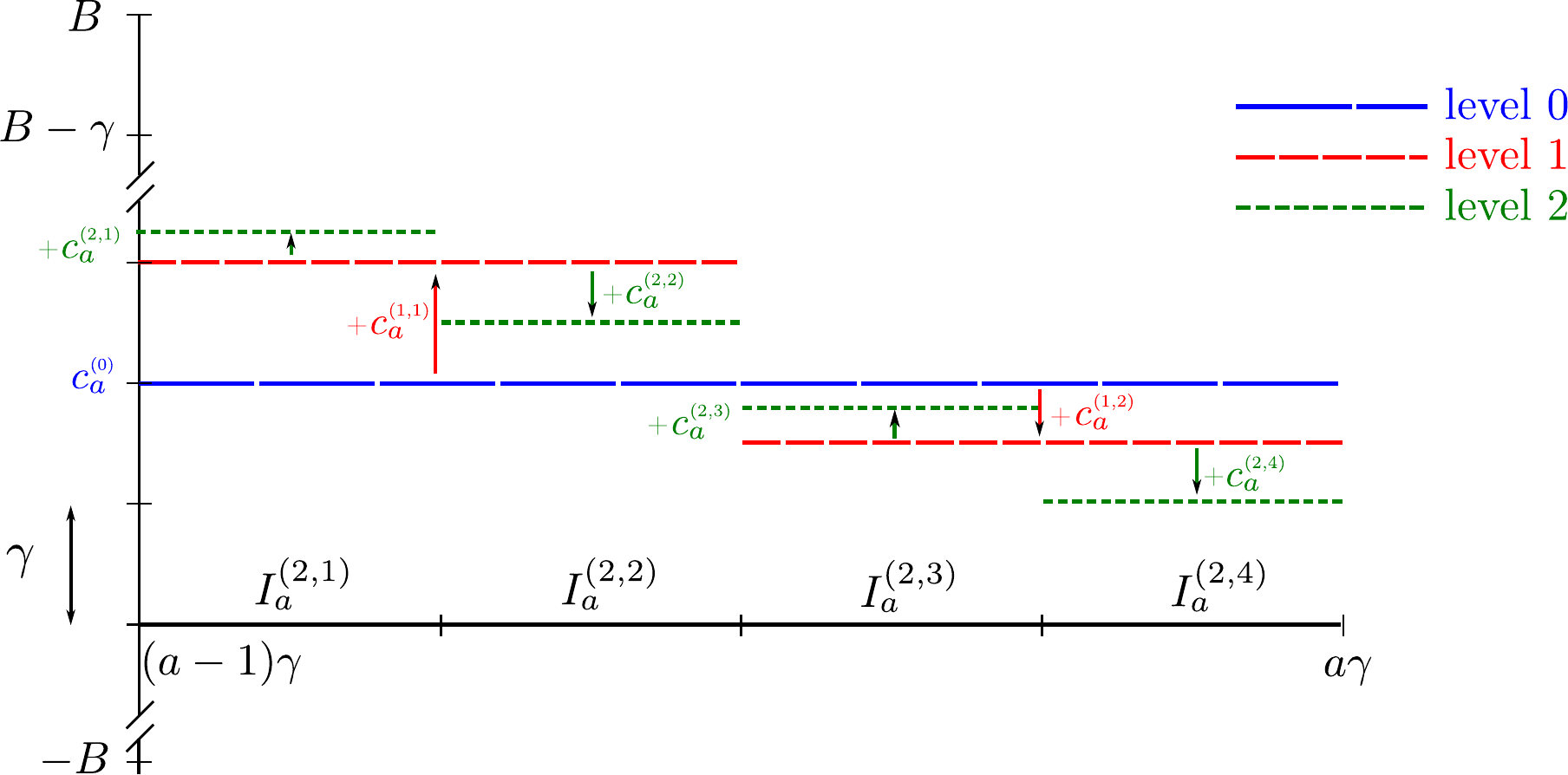}	
\caption{\label{fig:Lipschitz-deepernet} An example of function $f^{(0)}+\sum_{m=1}^M g^{(m)}$ for $M=2$, plotted on the subinterval $I_a$. This function corresponds to the dotted line (level $2$).}
\vspace{-1.5em}
\end{figure}

\paragraph{Refinement via a dyadic discretization} Next we construct $\gamma/2^m$-nets that are refinements of the $\gamma$-net $\cF^{(0)}$. We need to define a dyadic discretization for each subinterval $I_a$ as follows: for any level $m \geq 1$, we partition $I_a$ into $2^m$ subintervals $I_a^{(m,n)}$, $n=1,\ldots,2^m$, of equal size $\gamma/2^m$. Note that the subintervals $I_a^{(m,n)}$, $a=1,\ldots,1/\gamma$ and $n=1,\ldots,2^m$, form a partition of $[0,1]$. We call it \emph{the level-$m$ partition}. We enrich the set $\cF^{(0)}$ by looking at all the functions of the form $f^{(0)}+\sum_{m=1}^M g^{(m)}$, where $f^{(0)} \in \cF^{(0)}$ and where every function $g^{(m)}$ is piecewise-constant on the level-$m$ partition, with values $c^{(m,n)}_a \in \bigl[-\gamma/2^{m-1},\gamma/2^{m-1}\bigr]$ that are small when $m$ is large. In other words, we define \emph{the level-$M$ approximation set $\cF^{(M)}$} as the set of all functions $f_{\bc}:[0,1] \to \R$ of the form\\[-0.7cm]
\begin{align}
\hspace*{2.5cm} f_{\bc}(x) & = \underbrace{\sum_{a=1}^{1/\gamma} c_a^{(0)} \indicator{x \in I_a}}_{f^{(0)}(x)} + \sum_{m=1}^M \underbrace{\sum_{a=1}^{1/\gamma} \sum_{n=1}^{2^m} c^{(m,n)}_a \indicator{x \in I_a^{(m,n)}}}_{g^{(m)}(x)}~, \label{eq:Lip-def-FM}
\end{align}
\ \\[-0.2cm]
where $c_a^{(0)} \in \cC^{(0)}$ and $c^{(m,n)}_a \in \bigl[-\gamma/2^{m-1},\gamma/2^{m-1}\bigr]$. An example of function $f_{\bc} = f^{(0)}+\sum_{m=1}^M g^{(m)}$ is plotted on Figure~\ref{fig:Lipschitz-deepernet} in the case when $M=2$ (the plot is restricted to the interval $I_a$).\\[-0.2cm]

Since all functions in $\cF$ are $1$-Lipschitz, the set $\cF^{(M)}$ of all functions $f_{\bc}$ is a $\gamma/2^{M+1}$-net of $(\cF,\Vert\cdot\Vert_{\infty})$; see Lemma~\ref{lem:Lip-gammanets} in the appendix for a proof. Note that $\cF^{(M)}$ is infinite (the $c^{(m,n)}_a$ are continuously valued); fortunately this is not a problem since the $c^{(m,n)}_a$ can be rewritten as convex combinations $c^{(m,n)}_a = u^{(m,n)}_1 (-\gamma/2^{m-1}) + u^{(m,n)}_2 (\gamma/2^{m-1})$ of only two values; cf.~\eqref{eq:dyadicchaining-lossfunction} below.

%
%Using again the fact that all functions in $\cF$ are $1$-Lipschitz, we can show that the set $\cF^{(M)}$ of all functions $f_{\bc}$ is a $\gamma/2^{M+1}$-net of $(\cF,\Vert\cdot\Vert_{\infty})$; see Lemma~\ref{lem:Lip-gammanets} in the appendix for further details. Note that $\cF^{(M)}$ is infinite (the $c^{(m,n)}_a$ are continuously valued); fortunately this will not be a problem since the $c^{(m,n)}_a$ can be rewritten as convex combinations $c^{(m,n)}_a = u^{(m,n)}_1 (-\gamma/2^{m-1}) + u^{(m,n)}_2 (\gamma/2^{m-1})$ of only two values. We will use this fact right after~\eqref{eq:dyadicchaining-lossfunction} below.

%We can thus deduce as in~\eqref{eq:chainingEWA-regret-highscale-2} and~\eqref{eq:chainingEWA-regret-highscale-3} that, for $M \eqdef \bigl\lceil \log_2(\gamma T/B) \bigr\rceil$, we have
%\begin{equation}
%\inf_{\bc} \sum_{t=1}^T \bigl(y_t - f_{\bc}(x_t)\bigr)^2 \leq \inf_{f \in \cF} \sum_{t=1}^T \bigl(y_t - f(x_t)\bigr)^2 + \square B^2
%\label{eq:Lip-approximation}
%\end{equation}
%%
%for some absolute constant $\square$. Therefore, it is sufficient to be competitive against all functions $f_{\bc}$. This is what we do in the sequel.

\subsection{A chaining algorithm using this dyadic discretization}

Next we design an algorithm which, as in Section~\ref{sec:chainingEWA}, is able to be competitive against any function $f_{\bc} = f^{(0)}+\sum_{m=1}^M g^{(m)}$.
%A computationally naive approach would consist in maintaining a number of weights equal to $\card \cF^{(0)} + \sum_{m=1}^M \card \{g^{(m)\}$, which is infinite (the constants $c_a^{(m,n)}$ are continuously valued).
However, instead of maintaining exponentially many weights as in Algorithm~\ref{alg:chainingEWA}, we use the dyadic discretization in a crucial way. More precisely:
%\begin{itemize}
	%\item 
	
	We run $1/\gamma$ instances of the same algorithm $\cA$ in parallel; the $a$-th instance $\cA_a$, $a=1,\ldots,1/\gamma$, corresponds to the subinterval $I_a$ and it is updated only at rounds $t$ such that $x_t \in I_a$.
	%\item
	
	Next we focus on subalgorithm $\cA_a$.
	%whose local time is only incremented when a new $x_t$ falls into $I_a$.
As in Algorithm~\ref{alg:chainingEWA}, we use a combination of the EWA and the Multi-variable EG forecasters to perform high-scale and low-scale aggregation simultaneously:\\[0.2cm]
	\noindent
	\emph{Low-scale aggregation}: we run $2B/\gamma+1$ instances $\cB_{a,j}$, $j=0,\ldots,2B/\gamma$, of the Adaptive Multi-variable Exponentiated Gradient algorithm (Algorithm~\ref{alg:MultivarEG-adaptive} in the appendix) simultaneously. Each instance $\cB_{a,j}$ corresponds to a particular constant $c^{(0)}=-B+j\gamma \in \cC^{(0)}$ and is run (similarly to~\eqref{eq:chainingEWA-applicationMEG}) with the loss function $\ell_t$ defined for all weight vectors $\bu^{(m,n)}=\bigl(u^{(m,n)}_1,u^{(m,n)}_2\bigr) \in \Delta_2$ by
		\begin{align}
		& \ell_t\left(\bu^{(m,n)}, \, m =1,\ldots,M,\, n=1,\ldots,2^m\right) \nonumber \\
		& \quad = \left(y_t - (-B+j\gamma) - \sum_{m=1}^M \sum_{n=1}^{2^m} \left( u^{(m,n)}_1 \frac{-\gamma}{2^{m-1}} + u^{(m,n)}_2 \frac{\gamma}{2^{m-1}}\right) \indicator{x_t \in I_a^{(m,n)}} \right)^2~. \label{eq:dyadicchaining-lossfunction}
		\end{align}
	The above convex combinations $u^{(m,n)}_1 (-\gamma/2^{m-1}) + u^{(m,n)}_2 (\gamma/2^{m-1})$ ensure that subalgorithm~$\cB_{a,j}$ is competitive against the best constants $c^{(m,n)}_a \in \bigl[-\gamma/2^{m-1},\gamma/2^{m-1}\bigr]$ for all $m$ and $n$.\\[0.2cm]
	The weight vectors output by subalgorithm $\cB_{a,j}$ (when $x_t \in I_a$) are denoted by $\hat{\bu}^{(m,n)}_{t,a,j}$, and we set $\hat{f}_{t,a,j} (x) \!\eqdef \!-B+j\gamma + \sum_{m=1}^M \sum_{n=1}^{2^m} \!\left( \hat{u}^{(m,n)}_{t,a,j,1} \, \frac{-\gamma}{2^{m-1}} + \hat{u}^{(m,n)}_{t,a,j,2} \, \frac{\gamma}{2^{m-1}}\right) \indicator{x \in I_a^{(m,n)}}$ \mbox{for all $j=0,\ldots,2B/\gamma$.} \\[0.2cm]
\noindent
\emph{High-scale aggregation}: we aggregate the $2B/\gamma+1$ forecasters above with a standard Exponentially Weighted Average forecaster (tuned, e.g., with the parameter $\eta = 1/(2(4 B)^2)=1/(32B^2)$):\\[-0.2cm]
\begin{equation} \textstyle
\label{eq:dyadicchaining-EWA}
\hat{f}_{t,a} = \sum_{j=0}^{2B/\gamma} \hat{w}_{t,a,j} \hat{f}_{t,a,j}~.
\end{equation}
\noindent
\emph{Putting all things together}: at every time $t \geq 1$, we make the prediction $\hat{f}_t(x_t) \eqdef \sum_{a=1}^{1/\gamma} \hat{f}_{t,a}(x_t) \indicator{x_t \in I_a}$~. We call this algorithm the \emph{Dyadic Chaining Algorithm}. %\end{itemize}

\begin{theorem}
\label{thm:DyadicChaining}
Let $B>0$, $T \geq 2$, and $\cF$ be the set of all $1$-Lipschitz functions from $[0,1]$ to $[-B,B]$. Assume that $\max_{1 \leq t \leq T}|y_t| \leq B$. Then, the Dyadic Chaining Algorithm defined above and tuned with the parameters $\gamma=B T^{-1/3}$ and $M=\bigl\lceil \log_2(\gamma T/B) \bigr\rceil$ satisfies, for some absolute constant $c>0$, \\[-.7cm]
\[
\Reg_T(\cF) \leq c \max\{B,B^2\} T^{1/3} \log T~.
\]
\end{theorem}

\noindent
The proof is postponed to the appendix. Note that the Dyadic Chaining Algorithm is computationally tractable: at every round $t$, the  point $x_t$ only falls into one subinterval $I_a^{(m,n)}$ for each level $m=1,\ldots,M$, so that we only need to update $\mathcal{O}\bigl(2B/\gamma \times M\bigr) = \mathcal{O}\bigl(T^{1/3} \log T\bigr)$ weights at every round. For the same reason, the overall space complexity is $\mathcal{O}(T \times 2B/\gamma \times M) = \mathcal{O}\bigl( T^{4/3} \log T\bigr)$.

\begin{remark}
	The algorithm can be extended to the case of Lipschitz functions on $[0,1]^d$. It leads to an optimal regret of order $\cO(T^{d/(2+d)})$ up to a log factor. Besides, the computational complexity is still tractable. Indeed, at each round $t$, the point $x_t$ only falls into one cell of the partition. Hence, the time complexity is polynomial in $T$ with an exponent independent of $d$. This also applies to the space complexity if we use sparsity-tailored data types. 
	The extension to Hölder function classes on $[0,1]^d$ is however more difficult and we leave it for future work. 
\end{remark}

% Acknowledgments---Will not appear in anonymized version
\acks{The authors would like to thank Alexander S.\ Rakhlin for insightful initial discussions that triggered this work. The authors acknowledge the support of the French Agence Nationale de la Recherche (ANR), under grants ANR-13-BS01-0005 (project SPADRO) and  ANR-13-CORD-0020 (project ALICIA)}.

\bibliography{references-regressionalgo}

\appendix

\newpage
% ----------- ----------- ----------- ----------- ----------- ----------- ----------- -----------
\section{The chaining technique: a brief reminder}
% ----------- ----------- ----------- ----------- ----------- ----------- ----------- -----------
\label{sec:chaining-reminder}

The idea of chaining was introduced by~\citet{Dudley1967}. It provides a general method to bound  the supremum of stochastic processes. For the convenience of the reader, we recall the main ideas underlying this technique; see, e.g., \citet{BoLuMa-12-Concentration} for further details.
We consider a centered stochastic process $(X_f)_{f \in \cF}$ indexed by some finite metric space, say, $(\cF, \norm{\cdot}_{\infty})$, with subgaussian increments, which means that $\log \E e^{\lambda(X_f - X_{g})} \leq \frac{1}{2} v \lambda^2 \| f - g\|_{\infty}^2$ for all $\lambda > 0$ and all $f,g\in \cF$.
% that is,
%\[
	%\forall \lambda > 0, \quad \forall f,f'\in \cF \qquad \log \E e^{\lambda(X_f - X_{f'})} \leq \frac{1}{2} v \lambda^2 \| f - f'\|_{\infty}^2 \,.
%\]
%
\mbox{The goal is to bound the quantity $\E \big[ \sup_{f \in \cF} X_f \big] =  \E \big[ \sup_{f \in \cF} (X_f - X_{f_0})\big] $ for any $f_0 \in \cF$.}

\begin{lemma}[\citealt{BoLuMa-12-Concentration}] Let $Z_1,\dots,Z_K$ be subgaussian random variables with parameter $v > 0$  (i.e., $\log \E \exp(\lambda Z_i) \leq \lambda^2 v /2$ for all $\lambda \in \R$), then 
$
	\E \max_{i=1,\dots,K }Z_i \leq \sqrt{2v \log K} \,.
$
\label{lem:supremum-subgaussian}
\end{lemma}

\noindent
Lemma~\ref{lem:supremum-subgaussian} %(whose proof is provided in Appendix~\ref{sec:proofs}) 
entails
$
	\E \big[ \sup_{f \in \cF} (X_f-X_{f_0}) \big]  \leq  B \sqrt{2v  \log \left(\card \cF \right)} \,,
$
where $B = \sup_{f \in \cF} \| f - f_0\|_{\infty}$.
However, this bound is too crude since $X_f$ and $X_{g}$ are very correlated when $f$ and $g$ are very close. The chaining technique takes this remark into account by approximating the maximal value $\sup_f X_f$ by maxima over successive refining discretizations $\cF^{(0)},\dots,\cF^{(K)}$ of $\cF$. More formally, for any $f \in \cF$, we consider a sequence of approximations $\pi_0(f)=f_0 \in \cF^{(0)},\pi_1(f) \in \cF^{(1)},\dots,\pi_K(f)=f \in \cF^{(K)}$, where $\norm{f - \pi_k(f)}_{\infty} \leq B/2^k$ and $\card \cF^{(k)} = \cN_{\infty}(\cF,B /2^{k})$, so that:
\[
	\E \Big[ \sup_{f \in \cF} (X_f-X_{f_0}) \Big] =  \E \Bigg[ \sup_{f \in \cF} \sum_{k=0}^{K-1} \Big(X_{\pi_{k+1}(f)} - X_{\pi_{k}(f)} \Big)\Bigg] \leq \sum_{k=0}^{K-1} \E \Bigg[ \sup_{f \in \cF}  \Big(X_{\pi_{k+1}(f)} - X_{\pi_{k}(f)} \Big)\Bigg]  \,,
\]
We apply Lemma~\ref{lem:supremum-subgaussian} for each $k \in \{0, \dots,K-1\}$: since $\norm{\pi_{k+1}(f) - \pi_k(f)}_\infty \leq 3 B/2^{k+1}$ (by the triangle inequality) and $ \card \{ \pi_{k+1}(f) - \pi_k(f), f \in \cF\} \leq \cN_\infty(\cF, B/2^{k+1})^2$, we get the well-known Dudley entropy bound (note that $\epsilon \mapsto \cN_{\infty}(\cF,\epsilon)$ is nonincreasing):
\[
	\E \Big[ \sup_{f \in \cF} (X_f-X_{f_0})  \Big] \leq 6 \sum_{k=0}^{K-1} B2^{-k-1}\sqrt{v  \log \cN_\infty(\cF,B/2^{k+1}) } \leq 12  \sqrt{v} \int_0^{B/2} \sqrt{ \log \cN_\infty(\cF,\epsilon)} d\epsilon~.
\]
%where the last inequality is because $\epsilon \mapsto \cN_{\infty}(\cF,\epsilon)$ is nonincreasing.
%
%the variables $X_f - X_{f_0}$ are not correlated but extremely rude if they all are nearly equal. The idea of chaining consists in grouping nearly identical variables together before applying Lemma~\ref{lem:supremum-subgaussian}. To do so, one discretizes 

\newpage
% ----------- ----------- ----------- ----------- ----------- ----------- ----------- -----------
\section{Adaptive Multi-variable Exponentiated Gradient}
% ----------- ----------- ----------- ----------- ----------- ----------- ----------- -----------
\label{sec:MultivarEG-adaptive}

In this subsection, we provide an adaptive version of Algorithm~\ref{alg:MultivarEG} when the time horizon $T$ is not known in advance. We adopt the notations of Section~\ref{sec:MultivarEG}.
Basically, the fixed tuning parameters $\eta^{(1)},\dots,\eta^{(k)}$  are replaced with time-varying learning rates $\eta^{(1)}_t, \dots, \eta^{(k)}_t$.  

%Note also that to simplify the presentation, we only use constant tuning parameters $\eta^{(k)}$

\begin{algorithm2e}[h!]
\SetKwInOut{Input}{input}
\SetKwInOut{Init}{initialization}

\Input{optimization domain $\Delta_{N_1}\times \ldots \times \Delta_{N_K}$ (where $N_1,\ldots,N_K$ are positive integers).} 

\vspace{0.2cm}

\Init{set $\hat{\bu}^{(k)}_1 \eqdef \bigl(\frac{1}{N_k},\ldots,\frac{1}{N_k}\bigr) \in \Delta_{N_k}$ for all $k=1,\ldots,K$.}

\vspace{0.2cm}

\For{each round $t=1,2,\ldots$}{
\begin{itemize}
	\item Output $\bigl(\hat{\bu}^{(1)}_t,\ldots,\hat{\bu}^{(K)}_t\bigr) \in \Delta_{N_1}\times \ldots \times \Delta_{N_K}$ and observe the differentiable and jointly convex loss function $\ell_t:\Delta_{N_1}\times \ldots \times \Delta_{N_K} \to \R$.
	\item Update the tuning parameters, $\eta^{(k)}_{t}$ for all $k=1,\dots,K$ as follows:
		\[
			\eta^{(k)}_{t+1} = \frac{1}{G^{(k)}} \sqrt{\frac{\log N^{(k)}}{1+\sum_{s=1}^t \indicator{\norm{\nabla_{\bu^{(k)}} \ell_s}_\infty > 0}}}  
		\]
	\item Compute the new weight vectors $\bigl(\hat{\bu}^{(1)}_{t+1},\ldots,\hat{\bu}^{(K)}_{t+1}\bigr) \in \Delta_{N_1}\times \ldots \times \Delta_{N_K}$ as follows:
	\[
	\hat{\bu}^{(k)}_{t+1,i} \eqdef \frac{\exp\!\left(-\eta^{(k)}_{t+1} \displaystyle \sum_{s=1}^t \partial_{\hat{u}^{(k)}_{s,i}} \ell_s\!\left(\hat{\bu}^{(1)}_s,\ldots,\hat{\bu}^{(K)}_s\right)\right)}{Z^{(k)}_{t+1}}~, \quad i \in \{1,\ldots,N_k\},
	\]
	where $\partial_{\hat{u}^{(k)}_{s,i}} \ell_s$ denotes the partial derivative of $\ell_s$ with respect to $i$-th component of the vector variable~$\hat{\bu}^{(k)}_{s}$, and where the normalization factor $Z^{(k)}_{t+1}$ is defined by
	\[
	Z^{(k)}_{t+1} \eqdef \sum_{i=1}^{N_k} \exp\!\left(-\eta^{(k)}_{t+1} \sum_{s=1}^t \partial_{\hat{u}^{(k)}_{s,i}} \ell_s\!\left(\hat{\bu}^{(1)}_s,\ldots,\hat{\bu}^{(K)}_s\right)\right)~.
	\]
\end{itemize}
}
\caption{\label{alg:MultivarEG-adaptive}Adaptive Multi-variable Exponentiated Gradient}
\end{algorithm2e}

The Adaptive Multi-variable Exponentiated Gradient algorithm satisfies the regret bound of Theorem~\ref{thm:MultivarEG-adaptive} below.

\begin{theorem}
\label{thm:MultivarEG-adaptive}
Assume that the loss functions $\ell_t:\Delta_{N_1}\times \ldots \times \Delta_{N_K} \to \R$, $t \geq 1$, are differentiable and jointly convex. Assume also the following upper bound on their partial gradients: for all $k \in \{1,\ldots,K\}$,
\begin{equation}
\max_{1 \leq t \leq T} \norm{\nabla_{\bu^{(k)}} \ell_t}_{\infty} \leq G^{(k)}~.
\label{eq:MultivarEG-regret-assumptiongradients-adaptive}
\end{equation}
Then, the Multi-variable Exponentiated Gradient algorithm (Algorithm~\ref{alg:MultivarEG-adaptive}) has a regret bounded as follows:
\begin{align*}
\sum_{t=1}^T \ell_t\!\left(\hat{\bu}^{(1)}_t,\ldots,\hat{\bu}^{(K)}_t\right) 
	 - \min_{\bu^{(1)},\ldots,\bu^{(K)}} \sum_{t=1}^T \ell_t\!\left(\bu^{(1)},\ldots,\bu^{(K)}\right) 
	 \leq \; 2 \sum_{k=1}^K G^{(k)} \sqrt{ T^{(k)} \log N_k}~,
\end{align*}
where $T^{(k)} = \sum_{t=1}^T \indicator{\norm{\nabla_{\bu^{(k)}} \ell_t}_\infty > 0}$ and where the minimum is taken over all $\left(\bu^{(1)},\ldots,\bu^{(K)}\right) \in \Delta_{N_1}\times \ldots \times \Delta_{N_K}$.
\end{theorem}

\begin{proofref}{Theorem~\ref{thm:MultivarEG-adaptive}}
The proof starts as the one of Theorem~\ref{thm:MultivarEG}. From~\eqref{eq:MultivarEG-regret-1}, we can see that
\begin{align}
& \sum_{t=1}^T \ell_t\!\left(\hat{\bu}^{(1)}_t,\ldots,\hat{\bu}^{(K)}_t\right) - \min_{\bu^{(1)},\ldots,\bu^{(K)}} \sum_{t=1}^T \ell_t\!\left(\bu^{(1)},\ldots,\bu^{(K)}\right) \nonumber \\
& \qquad \qquad = \sum_{k=1}^K \left( \sum_{t=1}^T \bg^{(k)}_t \cdot \hat{\bu}^{(k)}_t - \min_{1 \leq i \leq N_k} \sum_{t=1}^T g^{(k)}_{t,i} \right) \nonumber \\
& \qquad \qquad = \sum_{k=1}^K \left( \sum_{t \in \cT^{(k)}} \bg^{(k)}_t \cdot \hat{\bu}^{(k)}_t - \min_{1 \leq i \leq N_k} \sum_{t \in \cT^{(k)}} g^{(k)}_{t,i} \right) ~, \label{eq:MultivarEG-regret-1-adaptive}
\end{align} 
where $\bg_t^{(k)} \eqdef \nabla_{\hat \bu^{(k)}_t} \ell_t(\hat \bu_t^{(1)}, \dots, \hat \bu_t^{(K)})$ and where $\cT^{(k)} = \big\{t =1,\dots,T,\quad \norm{\nabla_{\bu^{(k)}} \ell_t}_\infty > 0 \big\}$.\\

Note that the right-hand side of~\eqref{eq:MultivarEG-regret-1} is the sum of $K$ regrets. Let $k \in \{1,\ldots,K\}$. By definition of the Adaptive Multi-variable Exponentiated Gradient algorithm, the sequence of weight vectors $\bigl(\hat{\bu}^{(k)}_t\bigr)_{t \geq 1}$ corresponds exactly to the weight vectors output by the Exponentially Weighted Average forecaster with time-varying parameter (see Page~50 of \citealt{Ger-11-PhD}) applied to $N_k$ experts associated with the loss vectors $\bg^{(k)}_t \in \R^{N_k}$, $t \in \cT^{(k)}$. We can therefore use the well-known corresponding regret bound available, e.g., in Proposition~2.1 of \cite{Ger-11-PhD}. Noting that the loss vectors $\bg^{(k)}_t$ lie in $\bigl[-G^{(k)},G^{(k)}\bigr]^{N_k}$ by Assumption~\eqref{eq:MultivarEG-regret-assumptiongradients-adaptive}, and setting $T^{(k)} = \card \cT^{(k)}$, we thus get that
\begin{align*}
\sum_{t \in \cT^{(k)}} \bg^{(k)}_t \cdot \hat{\bu}^{(k)}_t - \min_{1 \leq i \leq N_k} \sum_{t \in \cT^{(k)}} g^{(k)}_{t,i} & \leq 2 G^{(k)} \sqrt{T^{(k)} \log N_k} \,. %+ G^{(k)} \sqrt{\log N_k}~. 
\end{align*}
Note that the additional term $G^{(k)} \sqrt{\log N_k}$ in the upper-bound of \cite{Ger-11-PhD} is actually not needed, since we can assume that $\eta^{(k)}_{T+1} = \eta^{(k)}_T$ because $\eta^{(k)}_{T+1}$ is not used by the algorithm at rounds $t \leq T$.
Substituting the last upper bound in the right-hand side of~\eqref{eq:MultivarEG-regret-1-adaptive} concludes the proof.
\end{proofref}

\newpage
\section{An efficient chaining algorithm for Hölder classes}
\label{sec:Holder-appendix}

In this appendix, we extend the analysis of Section~\ref{sec:Holder} to Hölder function classes. 
In the sequel $\cF$~denotes the set of functions on $[0,1]$ whose $q$ first derivatives ($q \in \N$) exist and are all bounded in supremum norm by a constant $B$, and whose $q$th derivative is Hölder continuous of order $\alpha \in (0,1]$ with coefficient $\lambda > 0$. 
In other words, any function $f \in \cF$ satisfies
\begin{equation}
	\label{eq:holder}
	\forall x,y \in [0,1], \quad \big|f^{(q)}(x) - f^{(q)}(y)\big| \leq \lambda |x-y|^\alpha \,,
\end{equation}
and $ \|{f^{(k)}}\|_{\infty} \leq B$ for all $k \in \{0,\dots,q\}$. 
We denote by $\beta = q+\alpha$ the coefficient of regularity of $\cF$. 
Recall from the introduction that $\log \cN_{\infty}(\cF,\epsilon) = \cO(\epsilon^{-1/\beta})$, so that, by Theorem~\ref{thm:chainingEWA-regret} and~\eqref{eq:DudleyRegretBound-csq}, if $\beta > 1/2$, the Chaining Exponentially Weighted Average forecaster guarantees a regret of $\cO \big(T^{1 / (2\beta+1)} \big)$, which is optimal.  We explain below how to modify this algorithm with non-proper $\epsilon$-nets of $(\cF, \norm{\cdot}_\infty)$ that are easier to manage from a computational viewpoint. This leads to a quasi-optimal regret of $\cO \big(T^{1 / (2\beta+1)} (\log T)^{3/2}\big)$.

\bigskip \noindent
The analysis follows the one of Section~\ref{sec:Holder} which  dealt with the special case of $1$-Lipschitz functions. The main difference consists in replacing piecewise-constant approximations with piecewise-polynomial approximations.

\subsection{Constructing computationally-manageable $\epsilon$-nets via exponentially nested discretization}

Let $\gamma \in \bigl(\frac{B}{T},B\bigr)$ be a fixed real number that will play the same role as in Theorem~\ref{thm:chainingEWA-regret}. Using the fact that all functions in $\cF$ are Hölder, we can approximate $\cF$ with piecewise-polynomial functions as follows. 

Let $\delta_x >0$ and $\delta_y > 0$ be two discretization widths that will be fixed later by the analysis.
We partition the $x$-axis $[0,1]$ into $1/\delta_x$ subintervals $I_a \eqdef \bigl[(a-1) \delta_x, a \delta_x \bigr)$, $a=1,\ldots,1/\delta_x$ (the last interval is closed at $x=1$). 
We also use a discretization of length $\delta_y$ on the $y$-axis $[-B,B]$, by considering the set 
	\[
		\cY^{(0)} \eqdef \Big\{-B+j \delta_y~:  \quad j=0,\ldots,2B/\delta_y \Big\} \,. 
	\]
For the sake of simplicity, we assume that both $1/\delta_x$ and $2B/\delta_y$ are integers. 
Otherwise, it suffices to consider $\lceil 1/\delta_x \rceil$ and $\lceil 2B/\delta_y \rceil$, which only impacts the constants of the final Theorem~\ref{thm:Hold-regret}. 
We then define the sets of clipped polynomial functions for every $a \in \{1,\dots,1/\delta_x\}$
	\[
		\cP_a^{(0)} \eqdef \left\{x \mapsto \bigg[a_0 + \frac{a_1}{1!} (x-x_a)^1 + \dots + \frac{a_q}{q!} (x-x_a)^q \bigg]_{B} ~: \quad  a_0,\dots,a_q \in \cY^{(0)} \right\}\,.
	\]
Here, $[\cdot]_{B}$ is the clipping operator defined by $[x]_{B} \eqdef \min \big\{B, \max\{-B,x\}\big\}$ and $x_a$ is the center of $I_a$. 
Now, we define the set $\cF^{(0)}$ of piecewise-clipped polynomial functions $f^{(0)}:[0,1] \to [-B,B]$ of the form
\begin{equation}
\label{eq:Hold-def-F0}
	\qquad f^{(0)}(x) = \sum_{a=1}^{1/\delta_x} P_a^{(0)}(x) \indicator{x \in I_a}~, \qquad \forall a \in \{1,\dots,1/\delta_x\} \quad P^{(0)}_a \in \cP_a^{(0)} ~.
\end{equation}
Remark that the above definition is similar to \eqref{eq:Lip-def-F0}, where the constants $c_a^{(0)}$ have been substituted with clipped polynomials. 
Using the fact that all functions in $\cF$ are Hölder, we can see (cf. Lemma~\ref{lem:Hold-gammanets}) that for $\delta_x = 2 (q! \gamma/(2\lambda))^{1/\beta}$ and $\delta_y = \gamma/e$,  the set $\cF^{(0)}$ is a $\gamma$-net\footnote{This $\gamma$-net is not proper since $\cF^{(0)} \not\subseteq \cF$.} of $\bigl(\cF,\Vert \cdot \Vert_{\infty} \bigr)$.

\paragraph{Refinement via an exponentially nested discretization} 
Next we construct $\gamma/2^m$-nets that are refinements of the $\gamma$-net $\cF^{(0)}$. 
We need to define an exponentially nested discretization for each subinterval $I_a$ as follows: 
for any level $m \geq 1$, we partition $I_a$ into $4^{m}$ subintervals $I_a^{(m,n)}$, $n=1,\ldots,4^{m}$, of equal size $\delta_x/4^{m}$. Note that the subintervals $I_a^{(m,n)}$, $a=1,\ldots,1/\delta_x$ and $n=1,\ldots,4^m$, form a partition of $[0,1]$. We call it \emph{the level-$m$ partition}. 

\medskip \noindent
Now, we design the sets of clipped polynomial functions $\cQ_a^{(m,n)}$ that will refine the approximation of $\cF$ on each interval $I_a^{(m,n)}$. To do so, for every $m \geq 1$ we set successive dyadic refining discretizations of the coefficients space $[-B,B]$:
	\begin{equation}
		\label{eq:def-cYm}
		\cY^{(m)} \eqdef \Big\{-B+j \delta_y / 2^m~:  \quad j=0,\ldots,2^{m+1}B/\delta_y \Big\} \,,
	\end{equation}
and we define the corresponding sets of clipped polynomial functions for all $a \in \{1,\dots,1/\delta_x\}$, all $m \in \{1,\dots, M\}$, and $n \in \{1,\dots,4^m\}$
	\begin{equation}
		\label{eq:def-cPamn}
		\cP_a^{(m,n)} \eqdef \left\{ x \mapsto \bigg[a_0 + \frac{a_1}{1!} \left(x-x_a^{(m,n)}\right)^1 + \dots + \frac{a_q}{q!} \left(x-x_a^{(m,n)}\right)^q \bigg]_{B}~: \quad a_0,\dots,a_q \in \cY^{(m)} \right\}\,,
	\end{equation}
where $x_a^{(m,n)}$ is the center of the interval $I_a^{(m,n)}$.
Then, we define the sets of differences between clipped polynomial functions of two consecutive levels
\[
	\cQ_a^{(m,n)} = \left\{ \Big[ P^{(m)} - P^{(m-1)} \Big]_{{3\gamma}/{2^m}}~: P^{(m)} \in \cP_a^{(m,n)} \mbox{ and } P^{(m-1)} \in \cP_a^{(m-1,n_{m-1})} 
		\right\} \,
\]
where $n_{m-1}$ denotes the  unique integer $n'$ such that $I_a^{(m,n)} \subset I_a^{(m-1,n')}$. (For $m=1$, $\cP_a^{(m-1,n_{m-1})} $ is replaced with $\cP_a^{(0)}$ in the definition of $\cQ_a^{(m,n)}$). The functions in $\cQ_a^{(m,n)}$ will play the same role as the constants $c_a^{(m,n)}$ for the Lipschitz case to refine the approximation from the level-$(m-1)$ partition to the level-$m$ partition. Note that each $Q_a^{(m,n)} \in \cQ_a^{(m,n)}$ takes values in $[-3\gamma/2^m,3\gamma/2^m]$.

\bigskip
Then, we enrich the set $\cF^{(0)}$ by looking at all the functions of the form $f^{(0)}+\sum_{m=1}^M g^{(m)}$, where $f^{(0)} \in \cF^{(0)}$ and where every function $g^{(m)}$ is the difference of a piecewise-clipped polynomial on the level-$m$ partition and a piecewise-clipped polynomial on the previous level $m-1$, with values $Q_a^{(m,n)} \in \cQ_a^{(m,n)}$.

\bigskip
In other words, we define \emph{the level-$M$ approximation set $\cF^{(M)}$} as the set of all functions $f_{\bc}:[0,1] \to \R$ of the form
\begin{align}
f_{\bc}(x) & = \underbrace{\sum_{a=1}^{1/\delta_x} P_a^{(0)}(x) \indicator{x \in I_a}}_{f^{(0)}(x)} + \sum_{m=1}^M \underbrace{\sum_{a=1}^{1/\delta_x} \sum_{n=1}^{4^m} Q_a^{(m,n)}(x) \indicator{x \in I_a^{(m,n)}}}_{g^{(m)}(x)}~, \label{eq:Hold-def-FM}
\end{align}
where $P_a^{(0)} \in \cP_a^{(0)}$ and $Q^{(m,n)}_a \in \cQ_a^{(m,n)}$. Once again, see \eqref{eq:Hold-def-FM} as an extension of \eqref{eq:Lip-def-FM}, where the constants $c_a^{(m,n)}$ have been replaced with $Q_a^{(m,n)}$.

\vspace{0.3cm}
Using again the fact that all functions in $\cF$ are Hölder, we can show that the set $\cF^{(M)}$ of all functions $f_{\bc}$ is a $\gamma/2^{M}$-net of $(\cF,\Vert\cdot\Vert_{\infty})$; see Lemma~\ref{lem:Hold-gammanets} below (whose proof is postponed to Appendix~\ref{sec:proofs-Hold-gammanets}) for further details.

\begin{lemma}
\label{lem:Hold-gammanets}
Let $\cF$ be the set of Hölder functions defined in~\eqref{eq:holder}. Assume that $\beta \eqdef q+\alpha \geq 1/2$. Let $\delta_x = 2 (q! \gamma/(2\lambda))^{1/\beta}$ and $\delta_y = \gamma/e$. Then: 
\begin{itemize}
	\item the set $\cF^{(0)}$ defined in~\eqref{eq:Hold-def-F0} is a $\gamma$-net of $\bigl(\cF,\Vert \cdot \Vert_{\infty} \bigr)$;
	\item for all $M \geq 1$, the set $\cF^{(M)}$ defined in~\eqref{eq:Hold-def-FM} is a $\gamma/2^{M}$-net of $\bigl(\cF,\Vert \cdot \Vert_{\infty} \bigr)$.
\end{itemize} 
\end{lemma}

\subsection{A chaining algorithm using this exponentially nested refining discretization}

Next we design an algorithm which, as in Section~\ref{sec:Holder}, is able to be competitive against any function $f_{\bc} = f^{(0)}+\sum_{m=1}^M g^{(m)}$ and is computationally tractable. More precisely:
	
	We run $1/\delta_x$ instances of the same algorithm $\cA$ in parallel; the $a$-th instance corresponds to the subinterval $I_a$ and it is updated only at rounds $t$ such that $x_t \in I_a$.
	
	Next we focus on the $a$-th instance of the algorithm $\cA$, whose local time is only incremented when a new $x_t$ falls into $I_a$. As in Algorithm~\ref{alg:chainingEWA}, we use a combination of the EWA and the Multi-variable EG forecasters to perform high-scale and low-scale aggregation simultaneously:\\[0.2cm]
	
	\noindent
	\emph{Low-scale aggregation}: we run $ \card \cP_a^{(0)} \leq (2B / \delta_y +1 )^{(q+1)}$ instances $\cB_{a,j}$, $j=1,\ldots, \card \cP_a^{(0)}$ of the Adaptive Multi-variable Exponentiated Gradient algorithm (Algorithm~\ref{alg:MultivarEG-adaptive} in the appendix) simultaneously. Each instance $\cB_{a,j}$ corresponds to a particular polynomial $P_{a,j}^{(0)} \in \cP_a^{(0)}$ and is run (similarly to \eqref{eq:chainingEWA-applicationMEG}) with the loss function $\ell_t$ defined for all weight vectors $\bu^{(m,n)} \in \Delta_{\card \cQ_a^{(m,n)}}$ by
		\begin{align}
		& \ell_t\left(\bu^{(m,n)}, \, m =1,\ldots,M,\, n=1,\ldots,4^m\right) \nonumber \\
		& \quad = \left(y_t - P_{a,j}^{(0)}(x_t) - \sum_{m=1}^M \sum_{n=1}^{4^m} \sum_{k=1}^{\ \card \cQ_a^{(m,n)}\!\!\!\!} u_{k}^{(m,n)} Q_{a,k}^{(m,n)}(x_t) \indicator{x_t \in I_a^{(m,n)}} \right)^2~. \label{eq:Hold-dyadicchaining-lossfunction}
		\end{align}
		Here, $Q_{a,1}^{(m,n)}, Q_{a,2}^{(m,n)},\dots$ denote the elements of $\cQ_a^{(m,n)}$ that have been ordered.
	The above convex combinations $\sum_{k} u_{k}^{(m,n)} Q_{a,k}^{(m,n)}$ ensure that subalgorithm $\cB_{a,j}$ is competitive against the best elements in $\cQ_a^{(m,n)}$ on subintervals $I_a^{(m,n)}$ for all $m$ and $n$.
	The weight vectors formed by this subalgorithm $\cB_{a,j}$ (when $x_t \in I_a$) are denoted by $\hat{\bu}^{(m,n)}_{t,a,j}$, and we set for all $j=1,\ldots, \card \cP_a^{(0)}$
	\[
		\hat{f}_{t,a,j} (x) \eqdef P^{(0)}_{a,j}(x) + \sum_{m=1}^M \sum_{n=1}^{4^m} \sum_{k=1}^{\ \card \cQ_a^{(m,n)}\!\!\!\!} \hat u_{t,a,j,k}^{(m,n)} \ Q_{a,k}^{(m,n)}(x) \indicator{x \in I_a^{(m,n)}} \,,
	 \]
	where $P_{a,j}^{(0)}$ is the $j$th element of $\cP_a^{(0)}$.\\[0.2cm]

\noindent
\emph{High-scale aggregation}: we aggregate the forecasters above $\hat f_{t,a,j}$ for $j \in \big\{1,\dots, \card \cP_a^{(0)} \big\}$ with a standard Exponentially Weighted Average forecaster (tuned, e.g., with the parameter $\eta = 1/(2(5 B)^2)=1/(50B^2)$):
\begin{equation}
\label{eq:Hold-dyadicchaining-EWA}
\hat{f}_{t,a} = \sum_{j=1}^{\card \cP_a^{(0)}} \hat{w}_{t,a,j} \hat{f}_{t,a,j}~.
\end{equation}
\noindent
\emph{Putting all things together}: at every time $t \geq 1$, we make the prediction $\hat{f}_t(x_t) \eqdef \sum_{a=1}^{1/\delta_x} \hat{f}_{t,a}(x_t) \indicator{x_t \in I_a}$~. We call this algorithm the \emph{Nested Chaining Algorithm for Hölder functions}. %\end{itemize}

\begin{theorem}
\label{thm:Hold-regret}
Let $B>0$, $T \geq 2$, and $\cF$ be the set of Hölder functions defined in~\eqref{eq:holder}. Assume that $\beta \eqdef q+\alpha \geq 1/2$ and that $\max_{1 \leq t \leq T}|y_t| \leq B$. Then, the Nested Chaining Algorithm for Hölder functions defined above and tuned with the parameters $\delta_x = 2 (q! \gamma/(2\lambda))^{1/\beta}$, $\delta_y = \gamma/e$, $\gamma= BT^{-\beta/(2\beta+1)}$ and $M=\bigl\lceil \log_2(\gamma T/B) \bigr\rceil$ satisfies, for some constant $c>0$ depending only on $q$ and $\lambda$,
\[
\Reg_T(\cF) \leq c \max\{B^{2-1/\beta},B^2\} T^\frac{1}{2\beta+1} (\log T)^{3/2}~.
\]
\end{theorem}

The proof is postponed to Appendix~\ref{sec:proofs-Hold-regret}. The logarithmic factor $(\log T)^{3/2}$ can be reduced to $\log T$, by partitioning $I_a$ into $2^{m/\beta}$ subintervals $I_a^{(m,n)}$ instead of $4^m$ subintervals. However, the partition at level $m \geq 2$ is then not necessarily nested in the partitions of lower levels, which makes the proof slightly more difficult. 

\bigskip \noindent
Note that the Nested Chaining Algorithm for Hölder functions is computationally tractable as shown by the following lemma, whose proof is deferred to Appendix~\ref{sec:proofs-Hold-complexity}.

\begin{lemma} \label{lem:Holder-complexity}
Under the assumptions of Theorem~\ref{thm:Hold-regret}, the complexity of the Nested Chaining Algorithm for Hölder functions defined above satisfies:
\begin{itemize}
	\item Storage complexity: $ \cO\Big(T^{2q+4+\frac{\beta (q-1)+1}{2\beta+1}} \log T\Big)$\,;
	\item Time complexity: $\cO \left( T^{(q+1)\left( 2+ \frac{\beta}{2\beta+1}\right)} \log T \right)$\,. 
\end{itemize}
\end{lemma}

\newpage
\section{Omitted proofs}

In this appendix, we provide the proofs which were omitted in the main body of the paper.

\subsection{Proof of Theorem~\ref{thm:MultivarEG}}
\label{sec:proofs-multivarEG}
As is the case for the classical Exponentiated Gradient algorithm, the proof relies on a linearization argument. Let $\left(\bu^{(1)},\ldots,\bu^{(K)}\right) \in \Delta_{N_1}\times \ldots \times \Delta_{N_K}$. By differentiability and joint convexity of $\ell_t$ for all $t=1,\ldots,T$, we have that
\begin{align}
& \sum_{t=1}^T \ell_t\!\left(\hat{\bu}^{(1)}_t,\ldots,\hat{\bu}^{(K)}_t\right) - \sum_{t=1}^T \ell_t\!\left(\bu^{(1)},\ldots,\bu^{(K)}\right) \nonumber \\
& \qquad \qquad \leq \sum_{t=1}^T \nabla \ell_t\!\left(\hat{\bu}^{(1)}_t,\ldots,\hat{\bu}^{(K)}_t\right) \cdot \left(\hat{\bu}^{(1)}_t-\bu^{(1)},\ldots,\hat{\bu}^{(K)}_t-\bu^{(K)}\right) \label{eq:MultivarEG-regret-01} \\
& \qquad \qquad = \sum_{t=1}^T \sum_{k=1}^K \nabla_{\hat{\bu}^{(k)}_t} \ell_t\!\left(\hat{\bu}^{(1)}_t,\ldots,\hat{\bu}^{(K)}_t\right) \cdot \left(\hat{\bu}^{(k)}_t-\bu^{(k)}\right)~, \label{eq:MultivarEG-regret-02}
\end{align}
where $\nabla \ell_t$ in \eqref{eq:MultivarEG-regret-01} denotes the usual (joint) gradient of $\ell_t$ (with $\sum_{k=1}^K N_k$ components), and where~\eqref{eq:MultivarEG-regret-02} follows from splitting the gradient into $K$ partial gradients: $\nabla \ell_t = \Bigl(\nabla_{\hat{\bu}^{(1)}_t} \ell_t,\ldots,\nabla_{\hat{\bu}^{(K)}_t} \ell_t\Bigr)$.\\

\noindent
As a consequence, setting $\bg^{(k)}_t \eqdef \nabla_{\hat{\bu}^{(k)}_t} \ell_t\!\left(\hat{\bu}^{(1)}_t,\ldots,\hat{\bu}^{(K)}_t\right) \in \R^{N_k}$, and taking the maximum of the last inequality over all $\left(\bu^{(1)},\ldots,\bu^{(K)}\right) \in \Delta_{N_1}\times \ldots \times \Delta_{N_K}$, we can see that
\begin{align}
& \sum_{t=1}^T \ell_t\!\left(\hat{\bu}^{(1)}_t,\ldots,\hat{\bu}^{(K)}_t\right) - \min_{\bu^{(1)},\ldots,\bu^{(K)}} \sum_{t=1}^T \ell_t\!\left(\bu^{(1)},\ldots,\bu^{(K)}\right) \nonumber \\
& \qquad \qquad \leq \sum_{k=1}^K \max_{\bu^{(k)} \in \Delta_{N_k}} \sum_{t=1}^T \bg^{(k)}_t \cdot \left(\hat{\bu}^{(k)}_t-\bu^{(k)}\right) \nonumber \\
& \qquad \qquad = \sum_{k=1}^K \left( \sum_{t=1}^T \bg^{(k)}_t \cdot \hat{\bu}^{(k)}_t - \min_{1 \leq i \leq N_k} \sum_{t=1}^T g^{(k)}_{t,i} \right)~, \label{eq:MultivarEG-regret-1}
\end{align}
where the last inequality follows from the fact that the function $\bu^{(k)} \mapsto \sum_{t=1}^T \bg^{(k)}_t \cdot \bu^{(k)}$ is linear over the polytope $\Delta_{N_k}$, so that its minimum is achieved on at least one of the $N_k$ vertices of $\Delta_{N_k}$.\\

Note that the right-hand side of~\eqref{eq:MultivarEG-regret-1} is the sum of $K$ regrets. Let $k \in \{1,\ldots,K\}$. By definition of the Multi-variable Exponentiated Gradient algorithm, the sequence of weight vectors $\bigl(\hat{\bu}^{(k)}_t\bigr)_{t \geq 1}$ corresponds exactly to the weight vectors output by the Exponentially Weighted Average forecaster (see Page~14 of \citealt{cesa-bianchi06prediction}) applied to $N_k$ experts associated with the loss vectors $\bg^{(k)}_t \in \R^{N_k}$, $t \geq 1$. We can therefore use the well-known corresponding regret bound available, e.g., in Theorem~2.2 of \cite{cesa-bianchi06prediction} or in Theorem~2.1 of \cite{Ger-11-PhD}. Noting that the loss vectors $\bg^{(k)}_t$ lie in $\bigl[-G^{(k)},G^{(k)}\bigr]^{N_k}$ by Assumption~\eqref{eq:MultivarEG-regret-assumptiongradients}, we thus get that
\begin{align*}
\sum_{t=1}^T \bg^{(k)}_t \cdot \hat{\bu}^{(k)}_t - \min_{1 \leq i \leq N_k} \sum_{t=1}^T g^{(k)}_{t,i} & \leq G^{(k)} \sqrt{2 T \log N_k}~.
\end{align*}
substituting the last upper bound in the right-hand side of~\eqref{eq:MultivarEG-regret-1} concludes the proof.

\subsection{An efficient $\gamma$-net for Lipschitz classes}

\begin{lemma}
\label{lem:Lip-gammanets} 
Let $\cF$ be the set of functions from $[0,1]$ to $[-B,B]$ that are $1$-Lipschitz. Then:
\begin{itemize}
	\item the set $\cF^{(0)}$ defined in~\eqref{eq:Lip-def-F0} is a $\gamma$-net of $\bigl(\cF,\Vert \cdot \Vert_{\infty} \bigr)$;
	\item for all $M \geq 1$, the set $\cF^{(M)}$ defined in~\eqref{eq:Lip-def-FM} is a $\gamma/2^{M+1}$-net of $\bigl(\cF,\Vert \cdot \Vert_{\infty} \bigr)$.
\end{itemize} 
\end{lemma}

\begin{proofref}{Lemma~\ref{lem:Lip-gammanets}} \\
\emph{First claim: $\cF^{(0)}$ is a $\gamma$-net of $\bigl(\cF,\Vert \cdot \Vert_{\infty} \bigr)$.} \\
Let $f \in \cF$. We explain why there exist $c^{(0)}_1,\ldots,c^{(0)}_{1/\gamma} \in C^{(0)}$ such that
\[
f^{(0)}(x) = \sum_{a=1}^{1/\gamma} c_a^{(0)} \indicator{x \in I_a}
\]
satisfies $\bigl|f(x)-f^{(0)}(x)\bigr| \leq \gamma$ for all $x \in [0,1]$. We can choose $c^{(0)}_a \in \argmin_{c \in C^{(0)}} \bigl|f(x_a)-c\bigr|$, where $x_a$ is the center of the subinterval~$I_a$. Indeed, since we can approximate $f(x_a)$ with precision $\gamma/2$ (the $y$-axis discretization is of width $\gamma$), and since $f$ is $1$-Lipschitz on $I_a$, we have that, for all $a \in \{1,\ldots,1/\gamma\}$ and all $x \in I_a$,
\[
\bigl|f(x) - c^{(0)}_a \bigr| \leq \bigl|f(x) - f(x_a) \bigr| + \bigl|f(x_a) - c^{(0)}_a \bigr| \leq \frac{\gamma}{2} + \frac{\gamma}{2} = \gamma~.
\]
Since the subintervals $I_a$, $a \leq 1/\gamma$, form a partition of $[0,1]$, we just showed that $\Vert f- f^{(0)}\Vert_{\infty} \leq \gamma$.

\ \\
\emph{Second claim: $\cF^{(M)}$ is a $\gamma/2^{M+1}$-net of $\bigl(\cF,\Vert \cdot \Vert_{\infty} \bigr)$.} \\
Let $f \in \cF$. We explain why there exist constants $c^{(0)}_a \in C^{(0)}$ and $c^{(m,n)}_a \in \bigl[-\gamma/2^{m-1},\gamma/2^{m-1}\bigr]$ such that
\[
f_{\bc}(x) = \sum_{a=1}^{1/\gamma} c_a^{(0)} \indicator{x \in I_a} + \sum_{m=1}^M \sum_{a=1}^{1/\gamma} \sum_{n=1}^{2^m} c^{(m,n)}_a \indicator{x \in I_a^{(m,n)}}
\]
satisfies $\bigl|f(x)-f_{\bc}(x)\bigr| \leq \gamma/2^{M+1}$ for all $x \in [0,1]$. We argue below that it suffices to:
\begin{itemize}
	\item choose the constants $c^{(0)}_a \in \argmin_{c \in C^{(0)}} \bigl|f(x_a)-c\bigr|$ exactly as for $\cF^{(0)}$ above;
	\item choose the constants $c^{(m,n)}_a$ in such a way that, for all levels $m \in \{1,\ldots,M\}$, and for all positions $a \in \{1,\ldots,1/\gamma\}$ and $n \in \{1,\ldots,2^m\}$,
	\begin{equation}
	f\bigl(x^{(m,n)}_a\bigr) = c_a^{(0)}+\sum_{m'=1}^m c^{(m',n_{m'})}_a~,
	\label{eq:Lip-deppernet-recursion}
	\end{equation}
	where $x^{(m,n)}_a$ denotes the center of the subinterval $I_a^{(m,n)}$, and where $n_{m'}$ is the unique integer $n'$ such that $I_a^{(m,n)} \subseteq I_a^{(m',n')}$. Such a choice can be done in a recursive way (induction on~$m$). It is feasible since the functions in $\cF$ are $1$-Lipschitz (see Figure~\ref{fig:Lipschitz-deepernet} for an illustration).
\end{itemize}

To conclude, it is now sufficient to use~\eqref{eq:Lip-deppernet-recursion} with $m=M$. Note indeed from~\eqref{eq:Lip-def-FM} that, on each level-$M$ subinterval~$I_a^{(M,n)}$, the function $f_{\bc}$ is equal to
\[
f_{\bc}(x) = c_a^{(0)} + \sum_{m=1}^M c^{(m,n_{m})}_a ~,
\]
where $n_{m}$ is the unique integer $n'$ such that $I_a^{(M,n)} \subseteq I_a^{(m,n')}$. Thus, by~\eqref{eq:Lip-deppernet-recursion}, we can see that $f_{\bc}\bigl(x^{(M,n)}_a\bigr) = f\bigl(x^{(M,n)}_a\bigr)$ for all points $x^{(M,n)}_a$, $a \in \{1,\ldots,1/\gamma\}$ and $n \in \{1,\ldots,2^M\}$.

Now, if $x \in I_a^{(M,n)}$ is any point in $I_a^{(M,n)}$, then it is at most at a distance of $\gamma/2^{M+1}$ of the middle point $x^{(M,n)}_a$. Therefore, by $1$-Lipschitzity of $f$, we have $\bigl|f(x) - f\bigl(x^{(M,n)}_a\bigr) \bigr| \leq \gamma/2^{M+1}$. Using the equality~$f_{\bc}\bigl(x^{(M,n)}_a\bigr) = f\bigl(x^{(M,n)}_a\bigr)$ proved above and the fact that $f_{\bc}$ is constant on $I_a^{(M,n)}$, we get that
\[
\forall a \in \{1,\ldots,1/\gamma\}, \quad \forall n \in \{1,\ldots,2^M\}, \quad \forall x \in I_a^{(M,n)}~, \qquad \bigl|f(x) - f_{\bc}(x)\bigr| \leq \gamma/2^{M+1}~.
\]
Since the level-$M$ subintervals $I_a^{(M,n)}$, $a \in \{1,\ldots,1/\gamma\}$ and $n \in \{1,\ldots,2^M\}$, form a partition of $[0,1]$, we just showed that $\Vert f - f_{\bc}\Vert_{\infty} \leq \gamma/2^{M+1}$, which concludes the proof.
\end{proofref}

\subsection{An efficient $\gamma$-net for Hölder classes (proof of Lemma~\ref{lem:Hold-gammanets})}
% -------------------------------------------------------------------

\label{sec:proofs-Hold-gammanets}

\emph{First claim: $\cF^{(0)}$ is a $\gamma$-net of $\bigl(\cF,\Vert \cdot \Vert_{\infty} \bigr)$.} \\
Let $f \in \cF$. We explain why there exist $P^{(0)}_a \in \cP_a^{(0)}$ for all $a \in \{1,\dots,1/\delta_x \}$ such that
\[
f^{(0)}(x) = \sum_{a=1}^{1/\delta_x} P_a^{(0)}(x) \indicator{x \in I_a}
\]
satisfies $\bigl|f(x)-f^{(0)}(x)\bigr| \leq \gamma$ for all $x \in [0,1]$. Fix $a \in \{1,\dots,1/\delta_x\}$ and let $x_a$ be the center of the subinterval~$I_a$. By Taylor's formula for all $x \in I_a$ there exist $\xi \in I_a$ such that
\begin{multline}
	f(x) = f(x_{a}) + f'(x_{a}) (x - x_{a})  + \frac{f''(x_{a})}{2!} (x-x_{a})^2 + \dots + \frac{f^{(q)}(x_{a})}{q!} (x - x_{a})^q \\
		+  \frac{1}{q!} \left(f^{(q)}(\xi) - f^{(q)}(x_{a}) \right)   (x - x_{a})^q\,.
		\label{eq:Taylor}
\end{multline}
Thus, the function $f$ can be written as the sum of a polynomial and a term (the last one) that will be proven to be small by the Hölder property~\eqref{eq:holder}. Now, for every derivative $i \in \{0,\dots,q\}$ we can choose $b_i \in \cY^{(0)}$ such that
\begin{equation}
	\label{eq:bi}
	|f^{(i)}(x_a) - b_i| \leq \delta_y / 2 \,.
\end{equation}
Indeed, the $y$-axis discretization $\cY^{(0)}$ of $[-B,B]$ is of width $\delta_y$ and $|f^{(i)}(x_a)| \leq B$ by definition of $\cF$. Thus, setting 
\[
	P_a^{(0)}(x) = b_0 + \frac{b_1}{1} (x-x_a) + \frac{b_2}{2!} (x-x_a)^2 + \dots + \frac{b_q}{q!} (x-x_a)^q \,,
\]
the polynomial $P_a^{(0)}$ satisfies by \eqref{eq:Taylor} for all $x \in I_a$
\begin{align*}
	\big|f(x) - P_a^{(0)}(x) \big| 
		& \leq \sum_{i=0}^q \frac{\big| f^{(i)}(x_a) -  b_i \big|}{i!} |x-x_a|^i + \frac{1}{q!} \left|f^{(q)}(\xi) - f^{(q)}(x_{a}) \right|   |x - x_{a}|^q \\
		&  \leq \sum_{i=0}^q \frac{\delta_y}{2i!} \underbrace{\bigg. |x-x_a|^i}_{\leq 1} + \frac{\lambda}{q!} \left|\xi - x_a \right|^\alpha   |x - x_{a}|^q \,,
\end{align*}
where the second inequality is by \eqref{eq:bi} and because $f^{(q)}$ is $\alpha$-Hölder with coefficient $\lambda$.
Now, since $|\xi - x_{a}|$ and  $|x - x_{a}|$ are bounded by $\delta_x /2$, it yields 
\[
	\big|f(x) - P_a^{(0)}(x) \big| \leq \sum_{i=0}^q \frac{\delta_y}{2i!}  +  \frac{\lambda}{q!}\left(\frac{\delta_x}{2}\right)^{q+\alpha}  \leq \frac{e}{2} \delta_y + \frac{\lambda}{q!}\left(\frac{\delta_x}{2}\right)^{\beta} \,.
\]
The choices $\delta_x = 2 (q! \gamma/(2\lambda))^{1/\beta}$ and $\delta_y = \gamma/e$ finally entail
\[
	 \Big|f(x) - \big[P_a^{(0)}(x)\big]_B \Big| \leq \big|f(x) - P_a^{(0)}(x)\big| \leq \frac{\gamma}{2} + \frac{\gamma}{2} = \gamma \,.
\]
This concludes the first part of the proof.

\ \\
\emph{Second claim: $\cF^{(M)}$ is a $\gamma/2^{M}$-net of $\bigl(\cF,\Vert \cdot \Vert_{\infty} \bigr)$.} \\
Let $f \in \cF$. We explain why there exist clipped-polynomials $P^{(0)}_a \in \cP^{(0)}$ and $Q^{(m,n)}_a \in \cQ_a^{(m,n)}$ such that
\begin{align*}
f_{\bc}(x) = \sum_{a=1}^{1/\delta_x} P_a^{(0)}(x) \indicator{x \in I_a}
	 + \sum_{m=1}^M \sum_{a=1}^{1/\delta_x} \sum_{n=1}^{4^m} Q_a^{(m,n)}(x) \indicator{x \in I_a^{(m,n)}}
\end{align*}
satisfies $\bigl|f(x)-f_{\bc}(x)\bigr| \leq \gamma/2^{M}$ for all $x \in [0,1]$. To do so, we show first that there exist clipped polynomials $P^{(0)}_a \in \cP^{(0)}$ and $P^{(m,n)}_a \in \cP_a^{(m,n)}$ such that
\begin{align*}
 \tilde f_{\bc}(x) = \sum_{a=1}^{1/\delta_x} P_a^{(0)}(x) \indicator{x \in I_a} & + \sum_{n=1}^{4} \sum_{a=1}^{1/\delta_x}  \big(P_a^{(1,n)}-P_a^{(0)}\big)(x) \indicator{x \in I_a^{(1,n)}}    \\
	& + \sum_{m=2}^M \sum_{a=1}^{1/\delta_x} \sum_{n=1}^{4^m} \Big( P^{(m,n)}_a - P^{(m-1,n_{m-1})}_a\Big)(x) \indicator{x \in I_a^{(m,n)}}
\end{align*}
satisfies $\bigl|f(x)-\tilde f_{\bc}(x)\bigr| \leq \gamma/2^{M}$ for all $x \in [0,1]$. We recall that $n_{m-1}$ denotes the  unique integer $n'$ such that $I_a^{(m,n)} \subset I_a^{(m-1,n')}$. First we remark that the function $\tilde f_c$  defined above equals $P_a^{(M,n)}$ on each level-$M$ subinterval $I_a^{(M,n)}$.

\bigskip \noindent
Thus, it suffices to design clipped polynomials $P_a^{(m,n)} \in \cP_a^{(m,n)}$, such that $\big|f(x) - P_a^{(m,n)}(x) \big| \leq \gamma / 2^m$ for all $x \in I_a^{(m,n)}$. To do so, we reproduce the same proof as for $\cF^{(0)}$ above. Because $\diam I_a^{(m,n)} = \delta_x /4^{m} \leq \delta_x / 2^{m/\beta}$ (recall that $\beta \geq 1/2$), for every position $a \in \{1,\dots,1/\delta_x\}$, every level $m \in \{1,\dots,M\}$, and every $n \in \{1,\dots, 4^m\}$, we can define as for $\cF^{(0)}$ above a polynomial 
\[
	\tilde P_a^{(m,n)}(x) = b_0 + \frac{b_1}{1} \left(x-x_a^{(m,n)}\right) + \frac{b_2}{2!} \left(x-x_a^{(m,n)}\right)^2 + \dots + \frac{b_q}{q!} \left(x-x_a^{(m,n)}\right)^q  \,
\]
(recall that $x_a^{(m,n)}$ is the center of $I_a^{(m,n)}$) such that all coefficients $b_j$ have the form $-B+ z_j \delta_y 2^{-m}$ for some $z_j \in \{0,\dots, 2^{m+1}B/\delta_y\}$ and 	\begin{equation}
		\label{eq:Qanm}
		\Big|f(x) - \big[ \tilde P_a^{(m,n)}(x)\big]_B \Big| \leq \gamma/2^m
	\end{equation}
for all $x \in I_a^{(m,n)}$. To conclude, we choose the clipped polynomials $P_a^{(m,n)} = \big[\tilde P_a^{(m,n)}\big]_B$. 

\bigskip \noindent
To conclude the proof, we see that for all $x \in I_a^{(m,n)}$, by the triangle inequality
\[
	 \left| P_a^{(m,n)}(x) - P_a^{(m-1,n_{m-1})}(x) \right| \leq \frac{\gamma}{2^m} + \frac{\gamma}{2^{m-1}} = \frac{3 \gamma}{2^m} \,,
\]
so that $f_c = \tilde f_c$ for the choices $Q_a^{(m,n)} = \Big[P_a^{(m,n)}(x) - P_a^{(m-1,n_{m-1})}(x)\Big]_{3\gamma/2^m}$.

\subsection{Proof of Theorem~\ref{thm:DyadicChaining}}
%------------------------------------------------------

\label{sec:proofs-DyadicChaining}

We split our proof into two main parts. First, we explain why each functions 
$\hat f_{t,a}$ incurs small cumulative regret inside each subinterval $I_a$. Second, we sum the previous regret bounds over all positions $a \in \{1,\dots,1/\gamma\}$.

\bigskip \noindent 
\paragraph{Part 1: focus on a subinterval $I_a$}
In this part, we fix some $a \in \{1,\dots,1/\gamma\}$ and we consider the $a$-th instance of the algorithm $\cA$, whose local time is only incremented when a new $x_t$ falls into $I_a$. As in Algorithm~\ref{alg:chainingEWA}, our instance of algorithm~$\cA$ uses a combination of the EWA and the Multi-variable EG forecasters to perform high-scale and low-scale aggregation simultaneously. Thus, the proof closely follows the path of the one of Theorem~\ref{thm:chainingEWA-regret}. We split again the proof into two subparts: one for each level of aggregation. 

\medskip \noindent
\emph{Subpart 1: low-scale aggregation.} \\[0.1cm]
In this subpart, we fix $j \in \{0,\ldots, 2B/\gamma\}$.   
The proof starts as the one of Theorem~\ref{thm:chainingEWA-regret} except that $\cA$ applies  the adaptive version of the Multi-variable Exponentiated Gradient forecaster (Algorithm~\ref{alg:MultivarEG-adaptive}, Appendix~\ref{sec:MultivarEG-adaptive}) with the loss function $\ell_t$ defined in~\eqref{eq:dyadicchaining-lossfunction}. % instead of Algorithm~\ref{alg:MultivarEG}. 
We will thus apply Theorem~\ref{thm:MultivarEG-adaptive} (available in Appendix~\ref{sec:MultivarEG-adaptive}) instead of Theorem~\ref{thm:MultivarEG}.
After checking its assumptions exactly as in the proof of Theorem~\ref{thm:chainingEWA-regret}, we can apply Theorem~\ref{thm:MultivarEG-adaptive}. The norms of the loss gradients $\big\| \nabla_{\hat u_t^{(m,n)}}\ell_t\big\|_{\infty} $ are bounded by $16B \gamma/2^m$ if $x_t$ falls in $I_a^{(m,n)}$ and by $0$ otherwise.
Setting $T^{(m,n)}_a = \sum_{t=1}^T \indicator{x_t \in I_a^{(m,n)}}$, Theorem~\ref{thm:MultivarEG-adaptive} yields as in \eqref{eq:chainingEWA-regret-lowscalecontribution-1}:
\begin{align}
	\sum_{t=1}^T & \left(y_t - \hat{f}_{t,a,j}(x_t)\right)^2  \indicator{x_t \in I_a}  \\
	& \quad \leq  \inf_{ c_a^{(m,n)}, \ \forall (m,n)}\ \  \sum_{t=1}^T \left(y_t - \biggl( -B+j\gamma  +  \sum_{m=1}^M \sum_{n=1}^{2^m} c_a^{(m,n)}   \indicator{x_t \in I_a^{(m,n)}} \biggr) \right)^2 \indicator{x_t \in I_a} \nonumber \\
		& \hspace*{3cm}  +  2 \sum_{m=1}^M \sum_{n=1}^{2^m} 16B \gamma /2^m  \sqrt{ T^{(m,n)}_a \log 2 }~,
\label{eq:chainingEWA-regret-lowscalecontribution-1-lipschitz}
\end{align}
where the infimum is over all constants $c_{a}^{(m,n)} \in [-\gamma/2^{m-1},\gamma/2^{m-1}]$ for every $m=1,\dots,M$ and $n=1,\dots,2^m$. But, for each level $m = 1,\dots, M$, the point $x_t$ only falls into one interval $I_a^{(m,n)}$. Thus, $\sum_{n=1}^{2^m} T^{(m,n)}_a = T_a$, where $T_a = \sum_{t=1}^T \indicator{x_t \in I_a}$ is the final local time of the $a$-th instance of $\cA$. Therefore, using the concavity of the square root and applying Jensen's inequality, \eqref{eq:chainingEWA-regret-lowscalecontribution-1-lipschitz} entails
\begin{align}
	\sum_{t=1}^T & \left(y_t - \hat{f}_{t,a,j}(x_t)\right)^2 \indicator{x_t \in I_a} \nonumber \\
		& \leq  \quad \inf_{ c_a^{(m,n)}, \ \forall (m,n)} \quad \sum_{t=1}^T \left(y_t - \bigg( -B+j\gamma  +  \sum_{(m,n) } c_a^{(m,n)} \indicator{x_t \in I_a^{(m,n)}} \bigg) \right)^2 \indicator{x_t \in I_a} \nonumber \\
	& \hspace*{4cm} + 32 B \gamma \sum_{m = 1}^M    2^{-m} \sqrt{T_a 2^m \log 2} \nonumber \\
		& \leq  \quad \inf_{ c_a^{(m,n)}, \ \forall (m,n)} \quad \sum_{t=1}^T \left(y_t - \bigg( -B+j\gamma  +  \sum_{(m,n) } c_a^{(m,n)} \indicator{x_t \in I_a^{(m,n)}} \bigg) \right)^2 \indicator{x_t \in I_a} \nonumber \\
	& \hspace*{4cm} + 32 B \gamma (1+\sqrt{2}) \sqrt{T_a \log 2} \,.
\label{eq:chainingEWA-regret-lowscalecontribution-3-lipschitz}
\end{align}
The second inequality is because $\sum_{m=1}^{\infty} 2^{-m/2} = 1 + \sqrt{2}$.

\bigskip \noindent
\emph{Subpart 2: high-scale aggregation}.\\[0.2cm]
Following the proof of Theorem~\ref{thm:chainingEWA-regret}, we apply EWA to the experts $\hat f_{t,a,j}$ for $j \in \{0,\dots,2B/\gamma\}$ with tuning parameter $\eta = 1/(2(4B)^2)$ because $\hat f_{t,a,j} \in [-B-2\gamma,B+2\gamma] \subset [-3B,3B]$ and $y_t \in [-B,B]$. We get from Proposition~3.1 and Page~46 of \cite{cesa-bianchi06prediction} that
\begin{align}
\sum_{t=1}^T & \left(y_t - \hat{f}_{t,a}(x_t)\right)^2 \indicator{x_t \in I_a} \nonumber \\
	& \leq  \min_{0 \leq j \leq 2B/\gamma} \quad \sum_{t=1}^T \left(y_t - \hat{f}_{t,a,j}(x_t)\right)^2 \indicator{x_t \in I_a} + \frac{\log \big( 2B/\gamma+1 \big) }{\eta} \nonumber \\
	&  \leq  \min_{0 \leq j \leq 2B/\gamma}  \quad \inf_{ c_a^{(m,n)},\, \forall (m,n)}	 \quad \sum_{t=1}^T \left(y_t - \bigg(-B+j\gamma  +  \sum_{(m,n)} c_a^{(m,n)} \indicator{x \in I_a^{(m,n)}}  \bigg)\right)^2 \indicator{x_t \in I_a} \nonumber \\
	& \hspace*{3cm} + 32  B \big(1+\sqrt{2}\big)\gamma \sqrt{T_a \log 2} \ + 32 B^2  \log \left( 2B/\gamma+1 \right)  ~, \label{eq:chainingEWA-regret-highscale-1-lipschitz}
\end{align}
where the infima are over all $j \in \{0,\dots,2B/\gamma\}$ and all constants $c^{(m,n)}_a \in [-\gamma/2^{m-1},\gamma/2^{m-1}]$, and where the second inequality follows from~\eqref{eq:chainingEWA-regret-lowscalecontribution-3-lipschitz} and from $\eta = 1/(32B^2)$. 

\paragraph{Part 2: we sum the regrets over all subintervals $I_a$} By definition of $\hat f_t$, we have
\begin{align*}
	\sum_{t=1}^T \big(y_t - \hat f_t(x_t)\big)^2 
		& = \sum_{t=1}^T \bigg(y_t - \sum_{a=1}^{1/\gamma} \hat f_{t,a}(x_t) \indicator{x_t \in I_a}\bigg)^2  \\
		& =\sum_{a=1}^{1/\gamma}  \sum_{t=1}^T  \bigg(y_t -  \hat f_{t,a}(x_t) \bigg)^2 \indicator{x_t \in I_a}
\end{align*}
Now, by definition of $\cF^{(M)}$, summing \eqref{eq:chainingEWA-regret-highscale-1-lipschitz} over all $a=1,\dots, 1/\gamma$ leads  to
\begin{align}
	\sum_{t=1}^T \big(y_t - \hat f_t(x_t)\big)^2 
		& \leq \inf_{f \in \cF^{(M)}} \sum_{t=1}^T  \big(y_t - f(x_t)\big)^2 + \frac{32 B^2}{\gamma}  \log \left( 2B/\gamma+1 \right) \nonumber \\
		& \hspace*{3cm} + 32  B \big(1+\sqrt{2}\big)\gamma \sqrt{\log 2} \left(\sum_{a=1}^{1/\gamma} \sqrt{T_a} \right) \   \,.
			\label{eq:jenaimarredechoisirdesrefs-lipschitz}
\end{align}
Then, using that $\sum_{a=1}^{1/\gamma} T_a = T$, since at every round $t$, the point $x_t$ only falls into one subinterval $I_a$, and applying Jensen's inequality to the square root, we can see that
\[
	\sum_{a=1}^{1/\gamma} \sqrt{T_a} \leq \sqrt{T/\gamma} \,.
\]
Therefore, substituting in~\eqref{eq:jenaimarredechoisirdesrefs-lipschitz}, we obtain
\begin{align}
	\sum_{t=1}^T \big(y_t - \hat f_t(x_t)\big)^2 
		& \leq \inf_{f \in \cF^{(M)}} \sum_{t=1}^T  \big(y_t - f(x_t)\big)^2 + \frac{32 B^2}{\gamma}  \log \left( 2B/\gamma+1 \right) \nonumber \\
		& \hspace*{3cm} + 32  B \big(1+\sqrt{2}\big) \sqrt{\gamma T \log 2  } \   \,.
		\label{eq:last-but-one-regret-lipschitz}
\end{align}
But, $\cF^{(M)}$ is by Lemma~\ref{lem:Lip-gammanets} a $\gamma/2^{M+1}$-net of $\cF$. Using that $M = \lceil \log_2(\gamma T/B) \rceil$ and following the proof of~\eqref{eq:chainingEWA-regret-highscale-3}, it entails
\[
	\inf_{f \in \cF^{(M)}} \  \sum_{t=1}^T  \big(y_t - f(x_t)\big)^2 \leq \inf_{f \in \cF}\  \sum_{t=1}^T  \big(y_t - f(x_t)\big)^2 + 2 B^2 + \frac{B^2}{4T} \,.
\]
Finally, from \eqref{eq:last-but-one-regret-lipschitz} we have
\begin{align}
	\sum_{t=1}^T \big(y_t - \hat f_t(x_t)\big)^2  
		& \leq \inf_{f \in \cF} \sum_{t=1}^T  \big(y_t - f(x_t)\big)^2 + \frac{32 B^2}{\gamma}  \log \left( 2B/\gamma+1 \right) \nonumber \\
		& \hspace*{3cm} + 32  B \big(1+\sqrt{2}\big) \sqrt{\gamma T \log 2  } + 2 B^2 + \frac{B^2}{4T} \   \,.
		\label{eq:last-regret-lipschitz}
\end{align}
The above regret bound grows roughly as (we omit logarithmic factors and small additive terms):
\[
	  \gamma^{-1} + \sqrt{\gamma T} \,.	
\]
Optimizing in $\gamma$ would yield $\gamma \approx T^{-1/3}$ and a regret roughly of the order of $T^{1/3}$. More rigorously, taking $\gamma=B T^{-1/3}$ and substituting it in~\eqref{eq:last-regret-lipschitz} concludes the proof.

\subsection{Proof of Theorem~\ref{thm:Hold-regret}}
% -------------------------------------------------------

\label{sec:proofs-Hold-regret}

The proof closely follows the one of Theorem~\ref{thm:DyadicChaining}.
It is split into two main parts. First, we explain why each function $\hat f_{t,a}$ incurs a small cumulative regret inside each subinterval $I_a$. Second, we sum the previous regret bounds over all positions $a=1,\dots,1/\delta_x$.

\bigskip \noindent 
\paragraph{Part 1: focus on a subinterval $I_a$}
In this part, we fix some $a \in \{1,\dots,1/\delta_x\}$ and we consider the $a$-th instance of the algorithm $\cA$, denoted $\cA_a$, whose local time is only incremented when a new $x_t$ falls into $I_a$. As in Algorithm~\ref{alg:chainingEWA}, $\cA_a$ uses a combination of the EWA and the Multi-variable EG forecasters to perform high-scale and low-scale aggregation simultaneously. 
We split again the proof into two subparts: one for each level of aggregation. 

\medskip \noindent
\emph{Subpart 1: low-scale aggregation.} \\[0.1cm]
In this subpart, we fix $j \in \{1,\ldots, \card \cP_a^{(0)}\}$. Similarly to the proof of Theorem~\ref{thm:DyadicChaining}, we start by applying Theorem~\ref{thm:MultivarEG-adaptive}. 
Since the elements in $\cQ_a^{(m,n)}$ are bounded in supremum norm by $3\gamma/2^m$, and since the elements in $\cP_a^{(0)}$ are bounded by $B$, the norms of the gradients of the loss function (defined in \eqref{eq:Hold-dyadicchaining-lossfunction}) are bounded by $0$ if $x_t \notin I_a^{(m,n)}$ and as follows otherwise:
\[
	\big\| \nabla_{\hat u_{t,a,j}^{(m,n)}}\ell_t\big\|_{\infty} \leq 2 \Big(|y_t| + \big\| \hat f_{t,a,j} \big\|_\infty \Big) \Big\| Q_{a,k}^{(m,n)} \Big\|_\infty \leq 2 ( B + 4B) 3\gamma/2^m = 30B \gamma/2^m\,.
\]
Here, we used that 
\begin{equation}
	\big| \hat f_{t,a,j}(x) \big|  \leq \big\|P_{a,j}^{(0)}\big\|_\infty + \sum_{m=1}^M \sum_{n=1}^{4^m} \sum_{k=1}^{\card \cQ_a^{(m,n)}} \hat u_{t,a,j,k}^{(m,n)} \Big|Q_{a,k}^{(m,n)}(x)\Big| \indicator{x \in I_{a}^{(m,n)}} \leq B + \sum_{m=1}^M \frac{3\gamma}{2^m} \leq 4B \,.
	\label{eq:bound-ftaj}
\end{equation}
Thus, setting $T^{(m,n)}_a = \sum_{t=1}^T \indicator{x_t \in I_a^{(m,n)}}$, Theorem~\ref{thm:MultivarEG-adaptive} yields:
\begin{align}
	\sum_{t=1}^T & \left(y_t - \hat{f}_{t,a,j}(x_t)\right)^2 \indicator{x_t \in I_a}  \nonumber \\
		& \leq  \inf_{ Q_a^{(m,n)}, \ \forall (m,n)}\ \   \sum_{t=1}^T \left(y_t - \left(P^{(0)}_{a,j}  +  \sum_{m=1}^M \sum_{n=1}^{4^m} Q_a^{(m,n)}   \indicator{x_t \in I_a^{(m,n)} }\right)(x_t)\right)^2 \indicator{x_t \in I_a} \nonumber \\
		& \hspace*{3cm}  +  2 \sum_{m=1}^M \sum_{n=1}^{4^m} 30B \gamma /2^m  \sqrt{ T^{(m,n)}_a \log \left( \card \cQ_a^{(m,n)} \right) }~,
\label{eq:chainingEWA-regret-lowscalecontribution-1-efficient}
\end{align}
where the infimum is over all polynomial functions $Q_{a}^{(m,n)} \in \cQ_a^{(m,n)}$ for every $m=1,\dots,M$ and $n=1,\dots,4^m$. But, for each level $m = 1,\dots, M$, the point $x_t$ only falls into one interval $I_a^{(m,n)}$. Thus, $\sum_{n=1}^{4^m} T^{(m,n)}_a = T_a$, where $T_a = \sum_{t=1}^T \indicator{x_t \in I_a}$ is the final local time of the $a$-th instance of $\cA$. Therefore, using the concavity of the square root and applying Jensen's inequality, \eqref{eq:chainingEWA-regret-lowscalecontribution-1-efficient} entails
\begin{align}
	\sum_{t=1}^T & \left(y_t - \hat{f}_{t,a,j}(x_t)\right)^2 \indicator{x_t \in I_a} \\
		& \leq  \inf_{ Q_a^{(m,n)}, \ \forall (m,n)} \sum_{t=1}^T \left(y_t - \left(P^{(0)}_{a,j}  +  \sum_{(m,n) } Q_a^{(m,n)} \indicator{x_t \in I_a^{(m,n)}}  \right)(x_t)\right)^2 \indicator{x_t \in I_a} \nonumber \\
	& \hspace*{3cm} + 60 B \gamma \sum_{m = 1}^M    2^{-m} \sqrt{T_a 4^m \log \left( \card \cQ_a^{(m,n)} \right)} \,.
\label{eq:chainingEWA-regret-lowscalecontribution-1bis-efficient}
\end{align}

\medskip \noindent
Now, by the definitions of $\cQ_a^{(m,n)}$, $\cP_a^{(m,n)}$, and $\cY^{(m)}$ (see Equations~\eqref{eq:def-cYm} and~\eqref{eq:def-cPamn}), we can see that
\begin{align*}
	\card \cQ_a^{(m,n)} & \leq \card \left(\cP_a^{(m,n)}\right)^2  
		 \leq  \Big( \card \cY^{(m)} \Big)^{2(q+1)} 
		 = \Big( 2^{m+1}B/\delta_y + 1 \Big)^{2(q+1)} \\
		& =  \Big( 2^{m+1}eB/\gamma + 1 \Big)^{2(q+1)} \,,
\end{align*}
which yields
\begin{align*}
	 \sum_{ m=1}^M {2^{-m}} \sqrt{4^m \log \left( \card \cQ_a^{(m,n)} \right)}
		& \leq  \sum_{m=1}^{M}  \sqrt{ 2 {(q+1)} \log  \big( 2^{m+1}eB/\gamma + 1\big)}  \\
		& \leq M \sqrt{2 {(q+1)} \log  \left( 2^{M+1}eB/\gamma + 1 \right)} \,.
\end{align*}
Thus, using $M= \lceil \log_2(\gamma T/B) \rceil$, so that $2^{M}  \gamma^{-1} \leq  2T/B$ and combining the above inequality with  \eqref{eq:chainingEWA-regret-lowscalecontribution-1bis-efficient}, we have
\begin{align}
	\sum_{t=1}^T & \left(y_t - \hat{f}_{t,a,j}(x_t)\right)^2 \indicator{x_t \in I_a} \\ 
	 &  \leq \quad \inf_{ Q_a^{(m,n)}, \ \forall (m,n)}	 \sum_{t=1}^T \left(y_t - \left(P^{(0)}_{a,j}  +  \sum_{(m,n)} Q_a^{(m,n)} \indicator{x_t \in I_a^{(m,n)}}  \right)(x_t)\right)^2 \indicator{x_t \in I_a} \nonumber \\
		& \hspace*{3cm} + 60  B \gamma  \lceil \log_2(\gamma T/B) \rceil
			\sqrt{2 (q+1) T_a \log(4eT+1)}  \,.
\label{eq:chainingEWA-regret-lowscalecontribution-3-efficient}
\end{align}\\[0.2cm]

\noindent \bigskip
\emph{Subpart 2: high-scale aggregation}.\\[0.2cm]
Following the proof of Theorem~\ref{thm:DyadicChaining}, we apply EWA to the experts $\hat f_{t,a,j}$ for $j \in \big\{1,\dots,\card \cP_a^{(0)} \big\}$ with tuning parameter $\eta = 1/(2(5B)^2) = 1/(50B^2)$ because $\hat f_{t,a,j} \in [-4B,4B]$ (see \eqref{eq:bound-ftaj}). From \eqref{eq:chainingEWA-regret-lowscalecontribution-3-efficient} and using $\card \cP_a^{(0)} \leq (2B/\delta_y + 1)^{q+1} = (2eB/\gamma + 1)^{q+1}$, we have 
\begin{align}
\sum_{t=1}^T & \left(y_t - \hat{f}_{t,a}(x_t)\right)^2 \indicator{x_t \in I_a} \nonumber \\
	& \leq  \min_{\ \ 1 \leq j \leq \card \cP_a^{(0)}} \quad \sum_{t=1}^T \left(y_t - \hat{f}_{t,a,j}(x_t)\right)^2 \indicator{x_t \in I_a} + \frac{\log \big( \card \cP_a^{(0)} \big) }{\eta} \nonumber \\
	&  \leq  \inf_{P^{(0)}_a \in \cP_a^{(0)}} \quad \inf_{ Q_a^{(m,n)}, \ \forall (m,n)}	 \quad \sum_{t=1}^T \left(y_t - \left(P^{(0)}_{a}  +  \sum_{(m,n)} Q_a^{(m,n)} \indicator{x_t \in I_a^{(m,n)}}  \right)(x_t)\right)^2 \indicator{x_t \in I_a} \nonumber \\
	& \hspace*{3cm} + 60  B \gamma 
			\lceil \log_2(\gamma T/B) \rceil 
			\sqrt{2(q+1) T_a\log(4eT+1)}  \nonumber \\
	& \hspace*{3cm} + 50 B^2 (q+1) \log \left( 2eB/\gamma +1\right)  ~, \label{eq:chainingEWA-regret-highscale-1-efficient}
\end{align}
where the infimum is over all functions $P^{(0)} \in \cP_a^{(0)}$ and $Q^{(m,n)}_a \in \cQ_a^{(m,n)}$, and where the second inequality follows from $\eta = 1/(50B^2)$.

\paragraph{Part 2: we sum the regrets over all subintervals $I_a$} By definition of $\hat f_t$, we have
\begin{align*}
	\sum_{t=1}^T \big(y_t - \hat f_t(x_t)\big)^2 
		& = \sum_{t=1}^T \bigg(y_t - \sum_{a=1}^{1/\delta_x} \hat f_{t,a}(x_t) \indicator{x_t \in I_a}\bigg)^2  \\
		& =\sum_{a=1}^{1/\delta_x}  \sum_{t=1}^T  \bigg(y_t -  \hat f_{t,a}(x_t) \bigg)^2 \indicator{x_t \in I_a}
\end{align*}
Now, by definition of $\cF^{(M)}$, summing \eqref{eq:chainingEWA-regret-highscale-1-efficient} over all $a=1,\dots, 1/\delta_x$ leads  to
\begin{align}
	\sum_{t=1}^T \big(y_t - \hat f_t(x_t)\big)^2 \leq \inf_{f \in \cF^{(M)}} & \sum_{t=1}^T  \big(y_t - f(x_t)\big)^2 \nonumber \\
		& + 50 B^2 (q+1) \log(2eB/\gamma + 1) \delta_x^{-1} \nonumber \\
		& + 60  B \gamma 
			\lceil \log_2(\gamma T/B) \rceil  
			\sqrt{2(q+1) \log(4eT+1)}  \left( \sum_{a=1}^{1/\delta_x} \sqrt{T_a} \right) \,.
			\label{eq:jenaimarredechoisirdesrefs}
\end{align}
Then, using that $\sum_{a=1}^{1/\delta_x} T_a = T$, since at every round $t$, the point $x_t$ only falls into one subinterval $I_a$, and applying Jensen's inequality to the square root, we can see that
\[
	\sum_{a=1}^{1/\delta_x} \sqrt{T_a} \leq \sqrt{T/\delta_x} \,.
\]
Therefore, substituting in~\eqref{eq:jenaimarredechoisirdesrefs} and because $\delta_x = 2 (q! \gamma/(2\lambda))^{1/\beta}$, we have
\begin{align*}
	\sum_{t=1}^T \big(y_t - \hat f_t(x_t)\big)^2 \leq \inf_{f \in \cF^{(M)}} & \sum_{t=1}^T  \big(y_t - f(x_t)\big)^2 \\
		& + 25 B^2 (q+1) \log(2eB/\gamma + 1)  \big(q! \gamma/(2\lambda)\big)^{-1/\beta} \\
		& + 60  B \gamma 
			\lceil \log_2(\gamma T/B) \rceil 
			\sqrt{(q+1) \log(4eT+1) T \big(q! \gamma/(2\lambda)\big)^{-1/\beta} } \,.
\end{align*}
But, $\cF^{(M)}$ is by Lemma~\ref{lem:Hold-gammanets} a $\gamma/2^M$-net of $\cF$. Using that $M = \lceil \log_2(\gamma T/B) \rceil$ and following the proof of~\eqref{eq:chainingEWA-regret-highscale-3}, it entails
\[
	\inf_{f \in \cF^{(M)}} \  \sum_{t=1}^T  \big(y_t - f(x_t)\big)^2 \leq \inf_{f \in \cF}\  \sum_{t=1}^T  \big(y_t - f(x_t)\big)^2 + 4 B^2 + \frac{B^2}{T} \,.
\]
Finally, we have
\begin{align}
	\sum_{t=1}^T \big(y_t - \hat f_t(x_t)\big)^2 \leq \inf_{f \in \cF} & \sum_{t=1}^T  \big(y_t - f(x_t)\big)^2 \nonumber \\
		& + 25 B^2 (q+1) \log(2eB/\gamma+1)  \big(q! \gamma/(2\lambda)\big)^{-1/\beta} + 4B^2 + \frac{B^2}{T} \nonumber \\
		& + 60  B \gamma 
			\lceil \log_2(\gamma T/B) \rceil 
			\sqrt{(q+1) \log(4eT+1) T \big(q! \gamma/(2\lambda)\big)^{-1/\beta} } \,.
			\label{eq:last-regret-Hold}
\end{align}
The above regret bound grows roughly as (we omit logarithmic factors and small additive terms):
\[
	  \gamma^{-1/\beta} + \gamma^{1 - 1/(2\beta)} \sqrt{T} \,.	
\]
Optimizing in $\gamma$ would yield $\gamma \approx T^{-\beta/(2\beta+1)}$ and a regret roughly of order $\cO\big(T^{1/(2\beta+1)}\big)$. More rigorously, taking $\gamma = B T^{-\beta/(2\beta+1)}$ and substituting it in~\eqref{eq:last-regret-Hold} concludes the proof.

\subsection{Proof of Lemma~\ref{lem:Holder-complexity}}
% ---------------------------------------------------------

\label{sec:proofs-Hold-complexity}

\paragraph{Storage complexity.} 
Fix a position $a \in \{1,\dots,1/\delta_x\}$.
At round $t \geq 1$, the Nested Chaining Algorithm for Hölder functions
needs to store:
\begin{itemize}
	\item the high-level weights $\hat w_{t,a,j}$ for every $j \in \big\{1,\dots, \card \cP_a^{(0)}\big\}$;
	\item the low-level weights 
$\hat u^{(m,n)}_{t,a,j,k}$ for every $j \in \big\{1,\dots,\card \cP_a^{(0)}\big\}$, every $m \in\{ 1,\dots,M \}$, every $n \in \{1,\dots,4^m\}$, and every $k \in \big\{1,\dots,\card \cQ_a^{(m,n)}\big\}$.
\end{itemize} 
The complexity of the $a$th instance of $\cA$ is thus of order
\[
	\card \cP_a^{(0)} \times M \times 4^M \times \card \cQ_a^{(M,n)} \,.
\]
Now, we bound each of these terms separately. First for $\gamma = BT^{-\beta/(2\beta+1)}$, we have 
\begin{align*}
	\card \cP_a^{(0)} 
		& \leq (2B/\delta_y+1)^{q+1} =(2eB/\gamma+1)^{q+1} = \big(2eT^{\beta/(2\beta+1)}+1\big)^{q+1} \\
		& = \cO\big(T^{\beta(q+1)/(2\beta+1)}) \,,
\end{align*}
because $\delta_y = e/\gamma$. Second using $M = \lceil \log_2(\gamma T /B)\rceil$, we can see that 
\[
	4^M = \big(2^M\big)^2 \leq (2\gamma T / B)^2 = \big(2 T^{1-\beta/(2\beta+1)}\big)^2 = \cO \big( T^{2-2\beta/(2\beta+1)}\big)  \,,
\]
and that
\[
	\card \cQ_a^{(M,n)} \leq \left(2^{M+1}eB/\gamma +1\right)^{2(q+1)} \leq \left(4 e T+1\right)^{2(q+1)} = \cO \big( T^{2(q+1)}\big) \,.
\]
Putting all things together the space-complexity of the $a$th instance of $\cA$ is of order
\[
	\cO \big( T^{2q+4+\beta(q-1)/(2\beta+1)} \log T\big)\,.
\]
The whole storage complexity of the algorithm is thus of order
\[
	\cO \big( T^{2q+4+\beta(q-1)/(2\beta+1)} / \delta_x \big) = \cO\big(T^{2q+4+(\beta (q-1)+1)/(2\beta+1)} \log T\big) \,,
\]
where we used that $\delta_x = 2 (q! \gamma/(2\lambda))^{1/\beta} = \cO\big( T^{-1/(2\beta+1)}\big)$

\bigskip \noindent
\textbf{Time complexity.} At round $t \geq 1$, $x_t$ only falls into one subinterval $I_a$ and one subinterval $I_a^{(m,n)}$ for each level $m=1,\dots,M$.
It thus needs to update 
\begin{itemize}
	\item the weights $\hat w_{t,a,j}$ for a single position $a$ and for every $j \in \big\{1,\dots,\card \cP_a^{(0)} \big\}$, 
	\item for every level $m=1,\dots,M$ the weights
$\hat u^{(m,n)}_{t,a,j,k}$ for a single position $a$ and a single $n$, but for all $j \in \big\{1,\dots,\card \cP_a^{(0)} \big\}$ and all $k \in \big\{1,\dots,\card \cQ_a^{(m,n)}\big\}$. 
\end{itemize}
The time-complexity is thus bounded by
\[
	\cO \Big(\card \cP_a^{(0)} \times  M  \times \card \cQ_a^{(M,n)}  \Big) = \cO \big( T^{(q+1) (2+ \beta/(2\beta+1)) } \log T \big) \,.
\]

\end{document}